\documentclass[a4paper,fleqn]{cas-dc}

\usepackage[authoryear,longnamesfirst]{natbib}

\usepackage{amsmath,booktabs,amssymb,bm,soul}

\usepackage{float}
\usepackage{hyperref}
\usepackage{subcaption}
\usepackage{caption}
\usepackage{bookmark}
\usepackage{epstopdf}
\usepackage{amsmath}

\usepackage{framed} 
\usepackage{multicol} 

\usepackage{nomencl} 
\def\tsc#1{\csdef{#1}{\textsc{\lowercase{#1}}\xspace}}
\tsc{WGM}
\tsc{QE}
\tsc{EP}
\tsc{PMS}
\tsc{BEC}
\tsc{DE}

\begin{document}
\let\WriteBookmarks\relax
\def\floatpagepagefraction{1}
\def\textpagefraction{.001}

\shorttitle{Two-dimensional TTC for articulated vehicles}

\shortauthors{Abhijeet Behera et~al.}

\title [mode = title]{Two-dimensional time-to-collision measures for articulated vehicles: predicting sideswipe and rear-end collisions }



%
\author[1,2]{Abhijeet Behera}[type=editor,
	auid=000,bioid=1,
	orcid=0009-0000-0356-7799]

\cormark[1]


\ead{abhijeet.behera@vti.se}


\credit{Conceptualization, Methodology, Formal analysis, Investigation, Writing-original draft}

\affiliation[1]{organization={ The Swedish National Road and Transport Research Institute},
    city={Linköping},
	postcode={58330},
	country={Sweden}}

\author[1,2]{Sogol Kharrazi}

\credit{Conceptualization, Writing - review \& editing, Supervision}

\author[2]{Erik Frisk}[
]
\credit{Resources, Writing - review \& editing, Supervision}

\author[1]{Maytheewat Aramrattana}
\credit{Funding acquisition, Writing - review \& editing}

\affiliation[2]{organization={Division of Vehicular Systems, Electrical Engineering, Linköping University},
	country={Sweden}}

\cortext[cor1]{Corresponding author}

\begin{abstract}
Time-to-collision is a commonly employed measure for rear-end collision prediction. However, its conventional formulation, which assumes constant speed and heading, is incapable of identifying sideswipe collisions. A two-dimensional extension has been proposed to incorporate lateral interactions with passenger cars, yet it assumes identical, fixed headings and does not accommodate articulated vehicles such as tractor–semitrailers. In this paper, the existing formulation for the car is first refined to incorporate differences in vehicle heading. Subsequently, new two-dimensional time-to-collision measures are proposed for articulated vehicles: TTC$^\text{AV}_\text{2D}$ and  modified TTC$^\text{AV}_\text{2D}$. These measures employ constant-speed and constant-acceleration assumptions and are analogous to their one-dimensional counterparts: time-to-collision and modified time-to-collision. The proposed measures are assessed in CARLA using randomly generated cut-in scenarios simulated with a tractor-semitrailer model, incorporating a range of trailer lengths. A short analysis is also conducted to test the measures in a roundabout and tight turn. \textcolor{black}{The analyses demonstrate that the proposed measures substantially improve the detection of sideswipe collisions while maintaining a comparable level of performance to existing measures in detecting rear-end collisions. Across $30$ simulated scenarios, they correctly identify $14$ of $15$ sideswipe collisions, compared with $7$ of $15$ identified by the existing formulation. Moreover, the mean prediction error for sideswipe collisions is reduced by approximately $20\%$ compared to the existing formulation.}
\end{abstract}


\maketitle

\begin{table*}
	\begin{framed}
		\begin{thenomenclature} 
\nomgroup{A}
  \item [{$\psi$}]\begingroup Yaw angle of the car\nomeqref {0}\nompageref{1}
  \item [{$\psi_0$}]\begingroup Yaw angle of the tractor\nomeqref {0}\nompageref{1}
  \item [{$\psi_1$}]\begingroup Yaw angle of the semitrailer\nomeqref {0}\nompageref{1}
  \item [{$a_X, a_Y$}]\begingroup Longitudinal and lateral accelerations of the car in its local coordinate system\nomeqref {0}\nompageref{1}
  \item [{$a_{0X}, a_{0Y}$}]\begingroup Transformed accelerations of the tractor in car’s local coordinate system\nomeqref {0}\nompageref{1}
  \item [{$a_{0x}, a_{0y}$}]\begingroup Longitudinal and lateral accelerations of the tractor in its local coordinate system\nomeqref {0}\nompageref{1}
  \item [{$c_{0X}, c_{0Y}$}]\begingroup Projected length and width of the tractor in the car's local coordinate system\nomeqref {0}\nompageref{1}
  \item [{$c_{1X}, c_{1Y}$}]\begingroup Projected length and width of the semitrailer in the car's local coordinate system\nomeqref {0}\nompageref{1}
  \item [{$l$}]\begingroup Length of the car\nomeqref {0}\nompageref{1}
  \item [{$l_{0}$}]\begingroup Length of the tractor\nomeqref {0}\nompageref{1}
  \item [{$l_{1}$}]\begingroup Length of the semitrailer\nomeqref {0}\nompageref{1}
  \item [{$l_{fa}, l_{ra}$}]\begingroup Distances from articulation joint to front of tractor and semitrailer\nomeqref {0}\nompageref{1}
  \item [{$R_0, R_1$}]\begingroup Rotation matrices from tractor and semitrailer to the car's local coordinate system\nomeqref {0}\nompageref{1}
  \item [{$s_{0X}, s_{0Y}$}]\begingroup Relative positions of the tractor with respect to the car in the $X$ and $Y$ directions\nomeqref {0}\nompageref{1}
  \item [{$s_{1X}, s_{1Y}$}]\begingroup Relative positions of the semitrailer with respect to the car in the $X$ and $Y$ directions\nomeqref {0}\nompageref{1}
  \item [{$v_X, v_Y$}]\begingroup Longitudinal and lateral speeds of the car in its local coordinate system\nomeqref {0}\nompageref{1}
  \item [{$v_{0X}, v_{0Y}$}]\begingroup Transformed speeds of the tractor in car’s local coordinate system\nomeqref {0}\nompageref{1}
  \item [{$v_{0x}, v_{0y}$}]\begingroup Longitudinal and lateral speeds of the tractor in its local coordinate system\nomeqref {0}\nompageref{1}
  \item [{$w$}]\begingroup Width of the car\nomeqref {0}\nompageref{1}
  \item [{$w_0$}]\begingroup Width of the tractor\nomeqref {0}\nompageref{1}
  \item [{$w_1$}]\begingroup Width of the semitrailer\nomeqref {0}\nompageref{1}
  \item [{$X, Y$}]\begingroup Longitudinal and lateral axes of the car's local coordinate system\nomeqref {0}\nompageref{1}
  \item [{$x, y$}]\begingroup Longitudinal and lateral axes of the tractor's local coordinate system\nomeqref {0}\nompageref{1}
  \item [{MTTC$^\text{AV}_\text{2D}$}]\begingroup Modified Two-dimensional time-to-collision for articulated vehicles\nomeqref {0}\nompageref{1}
  \item [{TTC}]\begingroup Time-to-collision\nomeqref {0}\nompageref{1}
  \item [{TTC$^\text{AV}_\text{2D}$}]\begingroup Two-dimensional time-to-collision for articulated vehicles\nomeqref {0}\nompageref{1}
  \item [{TTC$_\text{0X}$}]\begingroup Longitudinal TTC between the car and the tractor\nomeqref {0}\nompageref{1}
  \item [{TTC$_\text{0Y}$}]\begingroup Lateral TTC between the car and the tractor\nomeqref {0}\nompageref{1}
  \item [{TTC$_\text{1X}$}]\begingroup Longitudinal TTC between the car and the semitrailer\nomeqref {0}\nompageref{1}
  \item [{TTC$_\text{1Y}$}]\begingroup Lateral TTC between the car and the semitrailer\nomeqref {0}\nompageref{1}
  \item [{TTC$_\text{2D}$}]\begingroup Two-dimensional time-to-collision\nomeqref {0}\nompageref{1}

\end{thenomenclature}
	\end{framed}
\end{table*}

\section{Introduction}

Safety is a cornerstone of transportation planning and traffic management, impacting public health, infrastructure design, and economic efficiency. Traditional safety evaluation methods often rely on historical crash data to assess and mitigate risks \citep{abdel2007crash}. While these approaches provide valuable insights, they are inherently reactive, requiring accidents to occur before actionable conclusions can be drawn. This dependency on past incidents poses significant challenges, including ethical concerns, data limitations, and delays in addressing emerging safety issues \citep{tarko2018surrogate}. In this context, surrogate safety measures (SSMs) have emerged as a proactive and effective alternative for assessing traffic safety. It is a proxy that quantifies safety risks by measuring the likelihood of potential crashes. These measures rely on near-miss events, conflict analyses, and observable indicators, such as vehicle trajectories and traffic interactions, to predict and prevent accidents before they occur.

\begin{figure*}
	\centering
	\includegraphics[width=0.75\textwidth]{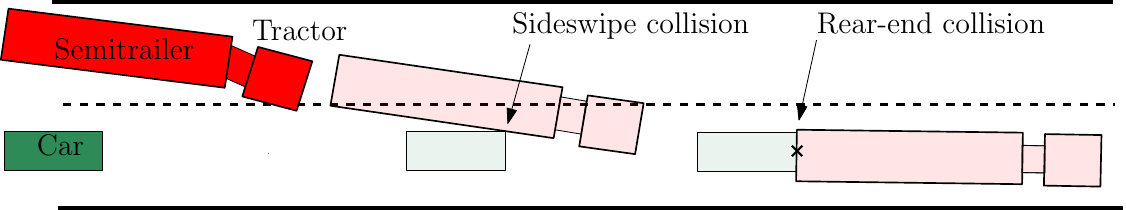}
	\caption{Collision between a tractor-semitrailer and a car during a lane change initiated by the tractor-semitrailer.}
	\label{fig:collision}
\end{figure*}

The inception of SSMs can be traced back to the early 1970s \citep{hayward1971near}, and considerable research has been done around it since then. Over the years, numerous SSMs have been developed to address a wide range of safety challenges. Researchers have adopted various approaches to categorise these SSMs. \cite{wang2021review} and \cite{mahmud2017application} provide a comprehensive review of SSMs and their categorisation. More specific analyses focusing on intersections and vulnerable road users are presented by \cite{sarkar2024review} and \cite{johnsson2018search}, respectively. Based on these studies, SSMs can be generally classified into four broad categories: Time, Distance, Deceleration, and Energy. Time-based SSMs include measures such as TTC and other variants of TTC developed over the years, post-encroachment time, time-to-accident, gap, headway \citep{hayward1971near,diwakar2024studying,behbahani2015new,do2025evaluation,nadimi2025assessing}. Distance-based SSMs comprise measures such as proportional stopping distance and unsafe density \citep{tak2018comparison,singh2025prisma}. Deceleration-based SSMs encompass measures such as deceleration rate to avoid collision, and crash potential index \citep{tak2015development,he2018assessing,fu2021comparison}. Energy-based SSMs involve measures such as crash severity, crash index, and deltaV \citep{shelby2011delta,stipancic2019modelling,hasanpour2025investigation}.

Time-to-Collision (TTC) is one of the most widely utilised surrogate safety measures (SSMs) in collision avoidance systems. It is primarily designed to predict rear-end collisions, where one vehicle impacts the rear of another \citep{hayward1972near}. \textcolor{black}{These collisions primarily arise from longitudinal interactions, where a following vehicle fails to maintain sufficient headway or respond in time to a decelerating lead vehicle. As a result, they are closely associated with car-following behaviour, reaction time, and braking performance, and can generally be described using one-dimensional motion along the direction of travel. In contrast, the conventional formulation of TTC does not account for the risk of lateral collisions, such as sideswipes, where contact occurs along the sides of the vehicles. These collisions are governed by relative lateral displacement, gap acceptance behaviour, and trajectory overlap rather than purely longitudinal distance. Consequently, modelling sideswipe conflicts requires a two-dimensional representation of vehicle motion.} For relatively short vehicles, such as passenger cars, the distinction between longitudinal and lateral collisions is less critical, since their dimensions in both directions are relatively similar. In such cases, even if a lateral collision occurs, but a longitudinal collision is predicted, the discrepancy in collision timing is typically small. However, this assumption does not hold for long articulated vehicles, such as tractor-semitrailers. Due to their significantly greater length compared to width, the physical dimensions introduce a large disparity between longitudinal and lateral collision risks. As a result, predicting collisions in only one direction (typically longitudinal) may overlook critical side-impact scenarios. Figure~\ref{fig:collision} illustrates two potential collision types that can arise when a tractor-semitrailer cuts into the path of a car.

It may be argued that preventing a rear-end collision would inherently eliminate the risk of a sideswipe collision in scenarios shown in Figure~\ref{fig:collision}. For instance, if a rear-end collision is anticipated, the tractor-semitrailer could avoid initiating the cut-in manoeuvre, thereby preventing both collisions. However, this approach results in an overly cautious driving style, which can lead to traffic flow disruptions \citep{haue2024automated}.

Several variants of TTC have been proposed, including time-exposed TTC, time-integrated TTC, and modified TTC \citep{minderhoud2001extended, ozbay2008derivation}. However, these formulations retain the core assumptions of the conventional TTC and are limited to longitudinal motion. To overcome this, \citet{hou2014new} introduced a geometric approach that considers both longitudinal and lateral directions by solving two-dimensional distance equations, while \citet{ward2015extending} proposed a looming-based method to evaluate collision risk in two dimensions. Building on these efforts, \citet{guo2023modeling} proposed a two-dimensional TTC (TTC$_\text{2D}$) that relaxes several simplifying assumptions of conventional TTC and quantifies collision risk in both longitudinal and lateral directions. The formulation assumes that both vehicles maintain identical and constant headings throughout the manoeuvre. While this assumption may be reasonable for gradual manoeuvres, such as long-distance highway lane changes, it becomes inadequate in scenarios involving substantial changes in heading angle, including short-distance lane changes, intersections, and roundabouts. These limitations become more pronounced in the case of articulated vehicles, such as a tractor and semitrailer, where the tractor and trailer may each have different heading angles. As a result, the applicability of TTC$_\text{2D}$ to a tractor-semitrailer remains unclear, and to date, no research has addressed two-dimensional collision-risk assessment for such vehicles. At the same time, advances in high-resolution sensors, cameras, and deep-learning–based perception systems now make it feasible to estimate the quantities required for two-dimensional collision-risk evaluation with high accuracy. This paper addresses these gaps by proposing new TTC$_\text{2D}$ measures for articulated vehicles.

To evaluate newly developed surrogate safety measures (SSMs), researchers typically adopt one of two primary approaches: analysis of historical crash data (e.g., \cite{nikolaou2023review, xie2019use}) or simulation-based evaluation (e.g., \cite{morando2018studying, so2015development, andreotti2025new, konrad2025cycling}). The traditional data-driven approach assesses the correlation between SSM outputs and observed crash frequencies or severities. While this method provides valuable empirical insights, it is often constrained by the availability and quality of crash data, and the time required to collect sufficient data, particularly for rare crash types \citep{he2018assessing}. Consequently, simulation-based methods that employ microscopic traffic environments, including SUMO and high-fidelity simulators such as CARLA, are gaining popularity (\cite{dosovitskiy2017carla, lopez2018microscopic}). In this paper, CARLA is employed to evaluate the proposed measures for articulated vehicles.

\textcolor{black}{The development of reliable safety measures for articulated vehicles is particularly important given their increasing role in freight transportation and their distinct operational characteristics compared with passenger vehicles. The larger dimensions, articulation dynamics, and manoeuvring constraints of these vehicles introduce safety challenges that are not fully captured by existing surrogate safety measures. Improving the representation of these interactions can support more accurate safety assessment in transportation simulation environments and provide a foundation for evaluating emerging vehicle technologies and operational strategies. Such advancements are also relevant for infrastructure planning and design, where accommodating diverse vehicle types requires a better understanding of vehicle-specific safety risks. By addressing these challenges, two-dimensional collision-risk measures for articulated vehicles can contribute to a more comprehensive approach to transportation safety analysis beyond conventional crash-based evaluations.}

\vspace{\baselineskip}
Therefore, the main contributions of this paper are summarised as follows:
\begin{itemize}
	\item \textcolor{black}{An extension of the existing two-dimensional TTC formulation to account for differences in vehicle heading through a coordinate transformation framework.}
    \begin{itemize}
        \item \textcolor{black}{The development of $\text{TTC}^{\text{AV}}_{\text{2D}}$, which is the first two-dimensional TTC measure specifically formulated for articulated vehicles and capable of predicting both rear-end and sideswipe collisions.}
        \item \textcolor{black}{The development of $\text{MTTC}^{\text{AV}}_{\text{2D}}$, a constant acceleration extension of $\text{TTC}^{\text{AV}}_{\text{2D}}$, analogous to the modified TTC concept but formulated in two dimensions for articulated vehicles.}
    \end{itemize}
	\item \textcolor{black}{An investigation of the proposed measures, $\text{TTC}^{\text{AV}}_{\text{2D}}$ and $\text{MTTC}^{\text{AV}}_{\text{2D}}$, using synthetic collision scenarios simulated in CARLA, followed by a comparative evaluation against existing surrogate safety measures.}
    
\end{itemize}

\section{\textcolor{black}{Existing TTC measures, limitations, and proposed improvements}} \label{sec:pd}

\textcolor{black}{This section reviews the conventional TTC and its extension in the literature to a two-dimensional TTC formulation. The limitations of both formulations are then discussed. Building on these observations, improvements are introduced to address the identified shortcomings. These improvements provide the foundation for the new formulation for articulated vehicles presented in Section \ref{sec:MF}.}

\subsection{Conventional TTC}
Conventional TTC is defined exclusively for longitudinal motion and represents the time required for a collision to occur between two vehicles, under the assumption that both maintain their current speeds and headings. The TTC is computed as:
\begin{equation}
	\text{TTC} =
	\begin{cases}
		\frac{s_{0\text{lon}} - \, l}{v_\text{lon} - v_{0\text{lon}}}, & v_\text{lon} > v_{0\text{lon}}, \\
		\infty,                                                        & \text{otherwise}.
	\end{cases}
	\label{eq:ttc_long}
\end{equation}
where $s_{0\text{lon}}$ is the longitudinal distance between the front ends of two vehicles, $v_{0\text{lon}}$  and $v_\text{lon}$ are the longitudinal speeds of the leading and following vehicles, respectively, and $l$ is the length of the leading vehicle.

\subsection{Existing Two-Dimensional TTC}

\begin{figure*}
	\centering
	\begin{subfigure}{0.49\textwidth}
		\centering
		\includegraphics[scale=0.65]{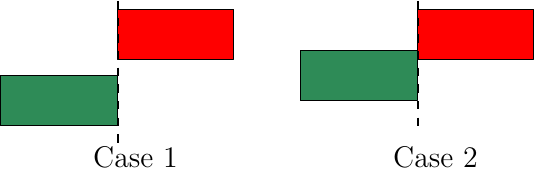}
		\caption{Rear-end collision.}
		\label{fig:RearEnd}
	\end{subfigure}
	\begin{subfigure}{0.49\textwidth}
		\centering
		\includegraphics[scale=0.65]{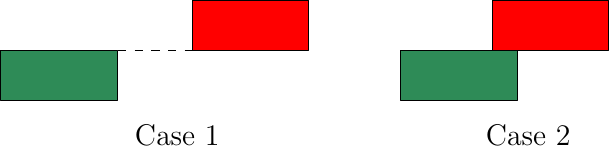}
		\caption{Sideswipe collision.}
		\label{fig:SideSwipe}
	\end{subfigure}
	\caption{Prospective collision scenarios. Red and green denote the lead and following cars, respectively.}
	\label{fig:collisionGuo}
\end{figure*}

\cite{guo2023modeling} extended the conventional definition of TTC by incorporating both longitudinal and lateral components. The resulting expression to compute  TTC$_\text{2D}$ is:
\begin{subequations}
\small
\begin{flalign}
\text{TTC}_{\text{lon}} &=
\begin{cases}
\dfrac{s_{0\text{lon}}-l}{v_{\text{lon}}-v_{0\text{lon}}},
&
\begin{aligned}
&s_{0\text{lon}}>l\;\&\; v_{\text{lon}}>v_{0\text{lon}}\\
&\&\;s_{0\text{lat}}
-(v_{\text{lat}}-v_{0\text{lat}})
\dfrac{s_{0\text{lon}}-l}{v_{\text{lon}}-v_{0\text{lon}}}
<w,
\end{aligned}
\\[-0.5ex]
\infty,&\text{otherwise.}
\end{cases}
&
\label{eq:GuoLong}
\\[1ex]
\text{TTC}_{\text{lat}} &=
\begin{cases}
\dfrac{s_{0\text{lat}}-w}{v_{\text{lat}}-v_{0\text{lat}}},
&
\begin{aligned}
&s_{0\text{lat}}>w\;\& \; v_{\text{lat}}>v_{0\text{lat}}\\
&\&\;s_{0\text{lon}}
-(v_{\text{lon}}-v_{0\text{lon}})
\dfrac{s_{0\text{lat}}-w}{v_{\text{lat}}-v_{0\text{lat}}}
<l,
\end{aligned}
\\[-0.5ex]
\infty,&\text{otherwise.}
\end{cases}
&
\label{eq:GuoLat}
\\[1ex]
\text{TTC}_{\text{2D}} &= 
\min\left(\text{TTC}_{\text{lon}},\text{TTC}_{\text{lat}}\right).
&
\end{flalign}
\label{eq:Guo}
\end{subequations}
where $\text{TTC}_\text{lon}$ is the time-to-collision in the longitudinal direction, and $\text{TTC}_\text{lat}$ is the time-to-collision in the lateral direction. $s_{0\text{lat}}$ is the initial lateral distance between the central longitudinal axes of the two vehicles, $v_{0\text{lat}}$  and $v_\text{lat}$ are the lateral speed of the leading and following vehicles, respectively, and $w$ is the width of the vehicle.

The first case outlined in (\ref{eq:GuoLong}) and (\ref{eq:GuoLat}) are based on three conditions: (a) There is no collision between the vehicles at the current time, (b) vehicles approach each other at a positive rate, (c) there is a lateral overlap at the predicted time of longitudinal collision, $\text{TTC}_\text{lon}$, and a longitudinal overlap at the time of lateral collision, $\text{TTC}_\text{lat}$. At the predicted time of collision in each direction, two possible scenarios can arise, as illustrated in Figure~\ref{fig:collisionGuo}. In Case 1, there is either longitudinal or lateral overlap, but not both. Hence, a collision does not occur. A collision occurs only in Case 2, where both longitudinal and lateral overlaps are present. The third condition in the first case of equations (\ref{eq:GuoLong}) and (\ref{eq:GuoLat}) ensures that Case 2 is obtained.

\subsection{Limitations of Existing TTC measures}

The conventional TTC defined in (\ref{eq:ttc_long}) is restricted to predicting rear-end collisions, as it accounts solely for longitudinal motion. Consequently, it neglects lateral vehicle interactions that are essential for capturing sideswipe collision risk. In addition, its application to articulated vehicles, such as tractor–semitrailers, remains non-trivial.

The two-dimensional TTC partially overcomes these limitations by incorporating both longitudinal and lateral motion components. Nevertheless, it introduces its own challenges. As implied by (\ref{eq:Guo}), the coordinate frames of the lead and following vehicles are assumed to be mutually aligned, a condition that rarely holds in real traffic environments, where vehicle orientations may differ substantially. Such misalignment influences both the estimated inter-vehicle distance and the relative speed vector. The problem becomes even more challenging for multi-unit articulated vehicles, as each unit can have a different orientation. Moreover, although the expressions in (\ref{eq:Guo}) can be applied straightforwardly to single-unit vehicles, they provide no clear mechanism for extension to articulated vehicles.

\subsection{Proposed improvement of two-dimensional TTC}

In this section, the baseline formulation in (\ref{eq:Guo}) is extended to account for vehicle orientation, assuming non-articulated vehicles. \textcolor{black}{Following the conventional TTC approach, speed and heading are assumed constant within the prediction horizon of a single time step. However, these parameters vary across the maneuver to account for changes in velocity and direction over time.}

\begin{figure*}
	\centering
	\includegraphics[width=0.5\textwidth]{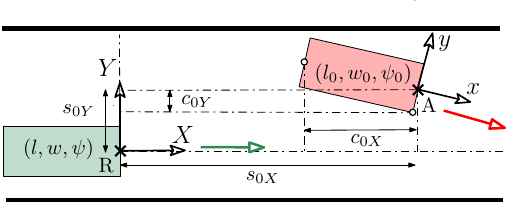}
	\caption{Schematic of scenario and employed coordinate system.}
	\label{fig:carCollision}
\end{figure*}

Figure~\ref{fig:carCollision} illustrates the coordinate system employed for calculating the relative position and speed of the vehicles involved. Local coordinate system, defined in accordance with the ISO vehicle system, are fixed to the following (green) vehicle at point $R$ and to the lead (red) vehicle at point $A$, both located at the midpoints of the front edges of the respective vehicles. The coordinate system of the following car is used as the Reference coordinate system ($R$). Thus, all computations are performed in the $R$-based coordinate system, using the transformed position and speed of the lead vehicle. This coordinate system comprises a longitudinal axis ($X$) and a lateral axis ($Y$). The quantities $s_{0X}$ and $s_{0Y}$ denote the relative position of point $A$ with respect to point $R$ along the $X$ and $Y$ axes, respectively.

In this formulation, the (lon, lat) representation used in the baseline model (see \ref{eq:Guo}) is replaced by the ($X$, $Y$) coordinate system. Let ($l_{0}, w_0, \psi_0$) represent the length, width, and yaw angle of the lead vehicle, and ($l, w, \psi$) refer to the following vehicle. To account for vehicle orientations, two changes are made to the baseline formulation in (\ref{eq:Guo}): (a) the effective projections of the lead vehicle's length and width onto the $X$ and $Y$ axes, denoted as $c_{0X}$ and $c_{0Y}$ respectively, are used in place of $l$ and $w$ in the baseline, (b) the longitudinal and lateral speeds of the lead vehicle, represented as $v_{0x}$ and $v_{0y}$ in its local coordinate system ($x, y$) are transformed in the $R$-based coordinate system, denoted as $v_{0X}$ and $v_{0Y}$. The projections are calculated as:
\begin{equation}
	\begin{bmatrix}
		c_{0X} \\
		c_{0Y}
	\end{bmatrix} = R_{0}
	\begin{bmatrix}
		l_{0} \\
		\frac{w_{0}}{2}
	\end{bmatrix},
	\label{eq:c0}
\end{equation}
where  $R_{0}$ is the rotation matrix is defined as
\begin{equation}
	R_{0} =
	\begin{bmatrix}
		\cos{(\psi_0 - \psi)} & -\sin{(\psi_0-\psi)} \\
		\sin{(\psi_0 - \psi)} & \cos{(\psi_0-\psi)}
	\end{bmatrix}.
	\label{eq:R02}
\end{equation}
Using the defined rotation matrix, the lead vehicle’s longitudinal and lateral speeds in the following vehicle’s coordinate system become:
\begin{equation}
	\begin{bmatrix}
		v_{0X} \\
		v_{0Y}
	\end{bmatrix} = R_{0}
	\begin{bmatrix}
		v_{0x} \\
		v_{0y}
	\end{bmatrix}.
	\label{eq:v0}
\end{equation}
The measure $\text{TTC}_\text{2D-Improved}$  is computed using the following expressions:
\begin{subequations}
\small
\begin{flalign}
& \text{TTC}_{0X} =
\begin{cases}
\dfrac{s_{0X}-c_{0X}}{v_X-v_{0X}},
&
\begin{aligned}
&s_{0X}>c_{0X}\;\&\; v_X>v_{0X}\\
&\&\;s_{0Y}-\frac{w}{2}
-(v_Y-v_{0Y}) \cdots \\
&\qquad\dfrac{s_{0X}-c_{0X}}{v_X-v_{0X}}
<c_{0Y},
\end{aligned}
\\[1ex]
\infty,&\text{otherwise.}
\end{cases}
\label{eq:ver1-X}
\\[1ex]
&\text{TTC}_{0Y} =
\begin{cases}
\dfrac{s_{0Y}-c_{0Y}-\frac{w}{2}}{v_Y-v_{0Y}},
&
\begin{aligned}
&s_{0Y}>c_{0Y}+\frac{w}{2}\;\&\; v_Y>v_{0Y}\\
&\&\;s_{0X}
-(v_X-v_{0X}) \cdots \\
&\qquad \dfrac{s_{0Y}-c_{0Y}-\frac{w}{2}}{v_Y-v_{0Y}}
<c_{0X},
\end{aligned}
\\[1ex]
\infty,&\text{otherwise.}
\end{cases}
\label{eq:ver1-Y}
\\[1ex]
& \text{TTC}_{\text{2D-Improved}}
=\min\left(\text{TTC}_{0X},\text{TTC}_{0Y}\right).
\label{eq:ver1}
\end{flalign}
\end{subequations}
Note that these expressions are similar to (\ref{eq:Guo}), but they take into account the projected lengths and widths, as well as the lead vehicle's longitudinal and lateral speeds, transformed into the coordinate system of the following vehicle.

\section{Extension of proposed two-dimensional TTC for articulated vehicles} \label{sec:MF}

In this section, the formulation derived in (\ref{eq:ver1}) is refined to accommodate articulated vehicles such as tractor–semitrailers. Two extensions are proposed. The first extension, denoted as $\text{TTC}_{\text{2D}}^{\text{AV}}$, assumes constant speed and heading over the prediction horizon. The second extension, denoted as modified $\text{TTC}_{\text{2D}}^{\text{AV}}$, which is abbreviated to $\text{MTTC}_{\text{2D}}^{\text{AV}}$, relaxes the assumption of constant speed in the prediction horizon and instead assumes constant acceleration. This formulation is analogous to the modified TTC (MTTC), a variant of the conventional TTC \citep{ozbay2008derivation}. The key difference is that MTTC operates in one dimension, whereas $\text{MTTC}_{\text{2D}}^{\text{AV}}$ operates in two dimensions. Note that the superscript `AV' indicates that the measure is specifically designed for articulated vehicles. Although real-world acceleration is not perfectly constant, this assumption is a reasonable approximation for heavy articulated vehicles due to their high inertia and low jerk.

\subsection{Tractor-semitrailer - car collision: constant speed and heading in the \textcolor{black}{prediction horizon}} \label{subsec:ver2}

Figure~\ref{fig:vehicleFrame1} shows the coordinate system employed for calculating the relative position of the tractor-semitrailer (red) with respect to the car (green). The specifics of the coordinate system and the vehicle dimensions are analogous to the previous section. The difference lies in the type of vehicles considered in this section. The tractor-semitrailer is an articulated vehicle, and hence, the articulation constraint is incorporated into the computation process. First, the measure $\text{TTC}_\text{2D,0}$ is computed for the collision between the tractor and the car. Since neither of them is an articulated vehicle, the computation process is similar to the previous section. After determining the $\text{TTC}_\text{2D,0}$ between the tractor and the car, a check is conducted to assess whether the semitrailer will collide with the car, given the tractor's trajectory. The measure $\text{TTC}_\text{2D,1}$ is computed between the semitrailer and the car. Subsequently, TTC$_\text{2D}^{\text{AV}}$ is determined as the minimum of the measures derived from individual units. This approach accounts for the physical coupling between the tractor and semitrailer, ensuring that the semitrailer's collision risk is evaluated based on the tractor’s trajectory and articulation angle.

\vspace{\baselineskip}

\subsubsection{TTC between the tractor and car}

The computation of $\text{TTC}_\text{2D,0}$ between the tractor and the car follows the same procedure as $\text{TTC}_\text{2D-Improved}$. Accordingly, the expressions derived in (\ref{eq:ver1}) remain applicable here.

\vspace{\baselineskip}
\subsubsection{TTC between the semitrailer and car}

A collision with the semitrailer can occur before a collision with the tractor. Given the trajectory of the tractor with respect to the car, the task is to identify the corresponding trajectory of the semitrailer with respect to the car. The forward euler discretisation method is employed to iteratively update the relative position of the tractor with respect to the car at each time step ($\Delta t$) until a collision occurs at time $\text{TTC}_\text{2D,0}$. The update equations are:
\begin{equation}
	\begin{aligned}
		s_{0X} \:(t) & =  s_{0X} \:(t-1) + (v_\text{X} - v_{0\text{X}}) \: \Delta t, \\
		s_{0Y} \:(t) & =  s_{0Y} \:(t-1) + (v_\text{Y} - v_{0\text{Y}}) \: \Delta t,
		\label{eq:sXY}
	\end{aligned}
\end{equation}
where the longitudinal ($v_{0X}$) and lateral ($v_{0Y}$) speeds are calculated in ($\ref{eq:v0}$). The heading of the tractor ($\psi_0$) stays constant for the entire computation duration, which starts from $t_0$, initial time, to $\text{TTC}_\text{2D,0}$, the estimated time for the collision between the tractor and the car. An analytical expression for the heading of semitrailer ($\psi_1$) is derived in (\ref{eq: psi1}) in Appendix. 

\begin{figure*}
	\centering
	\includegraphics[width=0.6\textwidth]{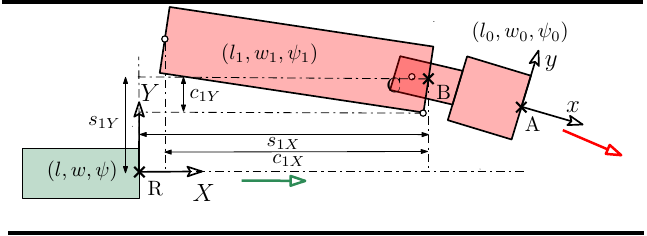}
	\caption{Schematic of scenario and employed coordinate system for the semitrailer of the articulated vehicle. The coordinate system for the tractor is the same as Figure~\ref{fig:carCollision}.}
	\label{fig:vehicleFrame1}%
\end{figure*}

Using the expressions for heading angles, the semitrailer's relative position with respect to the car ($s_{1X}, s_{1Y}$) at each time step is given by
\begin{equation}
	\begin{bmatrix}
		s_{1X} \:(t) \\
		s_{1Y} \:(t)
	\end{bmatrix} =
	\begin{bmatrix}
		s_{0X} \:(t) \\
		s_{0Y} \:(t)
	\end{bmatrix} -  R_0
	\begin{bmatrix}
		l_{fa} \\
		0
	\end{bmatrix} + R_1
	\begin{bmatrix}
		l_{ra} \\
		0
	\end{bmatrix},
	\label{eq:ver2_x2y2}
\end{equation}
where $l_{fa}$ is the distance of the tractor's front edge from the articulation joint, point $A$ to $C$ in Figure~\ref{fig:vehicleFrame1}, and $l_{ra}$ is the distance of the semitrailer's front edge from the articulation joint, point $B$ to $C$ in Figure~\ref{fig:vehicleFrame1}. The rotation matrix $R_{1}$ is defined as
\begin{equation}
	R_{1} =
	\begin{bmatrix}
		\cos{(\psi_1 - \psi)} & -\sin{(\psi_1-\psi)} \\
		\sin{(\psi_1 - \psi)} & \cos{(\psi_1-\psi)}
	\end{bmatrix}.
    \label{eq:r1comb}
\end{equation}
Let the projection of the semitrailer's length and width onto the $X$ and $Y$ axes be denoted as $c_{1X}$ and $c_{1Y}$:
\begin{equation}
	\begin{bmatrix}
		c_{1X} \\
		c_{1Y}
	\end{bmatrix} = R_{1}
	\begin{bmatrix}
		l_{1} \\
		\frac{w_{1}}{2}
	\end{bmatrix}.
	\label{eq:c1}
\end{equation}
The time-to-collision between the semitrailer and the car, denoted as $\text{TTC}_\text{2D,1}$, accounting for both longitudinal and lateral interactions, is determined by evaluating
\begin{subequations}
\small
\begin{align}
\text{TTC}_{1X} &=
\begin{cases}
t^*: &
\begin{aligned}
&s_{1X}(t_0)>c_{1X}(t_0)\;\&\\
&s_{1Y}(t^*)<c_{1Y}(t^*)+\frac{w}{2}\;\&\\
&s_{1X}(t^*)-c_{1X}(t^*)<\epsilon_X ,
\end{aligned}\\[1ex]
\infty: & \text{otherwise}.
\end{cases}
\label{eq: ver2a}
\\[1ex]
\text{TTC}_{1Y} &=
\begin{cases}
t^*: &
\begin{aligned}
&s_{1Y}(t_0)>c_{1Y}(t_0)+\frac{w}{2}\;\&\\
&s_{1X}(t^*)<c_{1X}(t^*)\;\&\\
&s_{1Y}(t^*)-c_{1Y}(t^*)-\frac{w}{2}<\epsilon_Y ,
\end{aligned}\\[1ex]
\infty: & \text{otherwise}.
\end{cases}
\label{eq: ver2b}
\\[1ex]
\text{TTC}_{\text{2D},1}
&=\min\left(\text{TTC}_{1X},\text{TTC}_{1Y}\right).
\label{eq: ver2c}
\end{align}
\label{eq:ver2}
\end{subequations}
where $t=t^*$ indicates the time stamp of the first contact (longitudinal or lateral) between the semitrailer and the car, $(\epsilon_{X},\:\epsilon_{Y})>0$ are small positive constants selected to account for numerical approximations in longitudinal and lateral contacts, respectively. The longitudinal and lateral collisions between the semitrailer and the car are checked at every time step in the prediction horizon using (\ref{eq: ver2a}) and (\ref{eq: ver2b}), respectively. Similar to  (\ref{eq:Guo}), there are three conditions for the first case in each of these expressions. The first condition ensures that there is no collision between the semitrailer and the car at the initial time $t=t_0$. The second and third conditions are satisfied only when there is a lateral overlap at the predicted time of longitudinal contact or a longitudinal overlap at the time of lateral contact, respectively.

Finally, the overall two-dimensional time-to-collision for the combined tractor–semitrailer system, denoted by $\text{TTC}_\text{2D}^{\text{AV}}$, is the minimum of the individual collision times:
\begin{equation}
	\text{TTC}_\text{2D}^\text{AV} = \min \:\left(\text{TTC}_\text{2D,0}, \: \text{TTC}_\text{2D,1}\right).
\end{equation}

\subsection{Tractor-semitrailer - car collision: constant acceleration and heading in the \textcolor{black}{prediction horizon}} \label{subsec:ver3}

In this extension, the assumption of constant speed in the prediction horizon is replaced with constant acceleration for both vehicles, while retaining the constant heading assumption. The calculation procedure follows the same approach as in the previous formulation, but now incorporates accelerations in both the longitudinal and lateral directions for the tractor-semitrailer and the car. As a result, some of the expressions derived in the second version are reformulated to include acceleration-based terms. In practice, it is not straightforward for the tractor-semitrailer to obtain the acceleration of the car. However, it is expected that with new technologies, it will become easy to estimate the acceleration of the surrounding vehicles. 

\subsubsection{TTC between the tractor and car}
Let $(a_X, a_Y)$ denote the longitudinal and lateral acceleration of the car in its local coordinate system ($X, Y$). The accelerations of the tractor,  transformed into the car's coordinate system, are denoted as ($a_{0X},  a_{0Y}$), and are given by
\begin{equation}
	\begin{bmatrix}
		a_{0X} \\
		a_{0Y}
	\end{bmatrix} = R_{0}
	\begin{bmatrix}
		a_{0x} - v_{0y} \dot{\psi}_0 \\
		a_{0y} + v_{0x} \dot{\psi}_0
	\end{bmatrix},
	\label{eq:a0}
\end{equation}
where $a_{0x}$ and $a_{0y}$ are the longitudinal and lateral speeds of the tractor in its local coordinate system ($x,y$) and $\dot{\psi}_0$ is the yaw rate of the tractor. Note that the heading of the tractor can change in real-time, but is constant in the prediction horizon. The same holds for the car. Since the car is travelling straight and the cut-in executed by the tractor-semitrailer, the yaw rate of the car and the corresponding Coriolis forces are assumed to be negligible.

Given the relative position, speed, and acceleration, the expressions that govern the tractor's and car's motion in both longitudinal and lateral directions are formulated. This leads to the following quadratic equations in time.
\begin{equation}
	\begin{aligned}
		s_{0X} & = (v_X - v_{0X})\:t_{0X} + 0.5\: (a_X-a_{0X})\:t_{0X}^2, \\
		s_{0Y} & = (v_Y - v_{0Y})\:t_{0Y} + 0.5\: (a_Y-a_{0Y})\:t_{0Y}^2.
	\end{aligned}
	\label{eq:position_v3_i}
\end{equation}
The roots of the equations determine the time required for the car to collide with the tractor, assuming both vehicles maintain their current headings. Each of the quadratic equations has two roots. Since the roots represent the time required for the collision, the smallest positive real root out of the two is selected. These roots, denoted as $t_{0X}^*$ for longitudinal and $t_{0Y}^*$ for lateral, are subsequently employed to compute the final relative position. The time-to-collision between the tractor and the car ($\text{TTC}_\text{2D,0}$) is then obtained using the following expressions: 

\begin{subequations}
\small
\begin{align}
\text{TTC}_{0X} &=
\begin{cases}
t_{0X}^{*}: &
\begin{aligned}
&s_{0X}>c_{0X}\;\&\\
&s_{0Y}-(v_Y-v_{0Y})\;t_{0X}^{*} \cdots \\
&\quad+0.5\;(a_Y-a_{0Y})\;{t_{0X}^{*}}^2 < c_{0Y}+\frac{w}{2},
\end{aligned}
\\[1ex]
\infty: &\text{otherwise}.
\end{cases}
\label{eq:3a}
\\[1ex]
\text{TTC}_{0Y} &=
\begin{cases}
t_{0Y}^{*}: &
\begin{aligned}
&s_{0Y}>c_{0Y}+\frac{w}{2}\;\&\\
&s_{0X}-(v_X-v_{0X})\;t_{0Y}^{*} \cdots\\
&\quad +0.5\;(a_X-a_{0X})\;{t_{0Y}^{*}}^2 <c_{0X},\\
\end{aligned}
\\[1ex]
\infty: &\text{otherwise}.
\end{cases}
\label{eq:3b}
\\[1ex]
\text{TTC}_{\text{2D},0}
&=\min\left(\text{TTC}_{0X},\text{TTC}_{0Y}\right).
\label{eq:3c}
\end{align}
\label{eq:ver3}
\end{subequations}
In contrast to (\ref{eq:ver1}), there are two conditions for the first case in (\ref{eq:3a}) and (\ref{eq:3b}). Since the vehicles are no longer moving at constant speed, the speed constraints do not hold and are replaced by new constraints on the existence of a positive root for the quadratic equation (\ref{eq:position_v3_i}). The first condition checks that there is no collision between the vehicles at the current time. In the second condition, the overlap between the vehicles is computed using constant acceleration instead of constant speed.

\subsubsection{TTC between the semitrailer and car}

There are only two changes compared to the first extension. The rest of the expressions stay the same. First, the expressions in (\ref{eq:sXY}) are reformulated to include the acceleration terms. The modified expressions are:
\begingroup
\small
\begin{equation}
\begin{aligned}
s_{0X}(t)
&= s_{0X}(t-1)
+(v_X-v_{0X})\,\Delta t 
+0.5\,(a_X-a_{0X})\,(\Delta t)^2,
\\[1ex]
s_{0Y}(t)
&= s_{0Y}(t-1)
+(v_Y-v_{0Y})\,\Delta t 
+0.5\,(a_Y-a_{0Y})\,(\Delta t)^2.
\end{aligned}
\label{eq:update}
\end{equation}
\endgroup
The second modification involves using the expression derived in (\ref{eq: psi1_acc}) in the Appendix for the semitrailer's heading angle, instead of (\ref{eq: psi1}). This updated expression is incorporated into the expressions presented in (\ref{eq:ver2_x2y2}) and (\ref{eq:c1}).

The final set of expressions for calculating the modified time-to-collision between the semitrailer and the car \(\text{TTC}_\text{2D,1}\) are the same as (\ref{eq:ver2}). Consequently, the measure \(\text{MTTC}_\text{2D}^{\text{AV}}\) is defined as:
\begin{equation}
	\text{MTTC}_\text{2D}^\text{AV} = \min \left( \text{TTC}_\text{2D,0}, \text{TTC}_\text{2D,1} \right).
\end{equation}

The proposed measures are formulated for scenarios where the car approaches the tractor-semitrailer from behind. However, they can be easily adapted for cases where the tractor-semitrailer approaches the car from behind. This adaptation requires replacing the tractor or semitrailer length with the car's length in all relevant expressions. Additionally, the relative longitudinal speed of the car with respect to the tractor-semitrailer must be negative to ensure a positive approach rate of the tractor-semitrailer towards the car.

\textcolor{black}{The proposed measures can be used for articulated vehicles with more than two units.} For such cases, the computation process follows the same logic as that used for the semitrailer in this study. For each unit after the first, collision conditions are evaluated up to the time-to-collision of the preceding unit. The overall time-to-collision for the articulated vehicle is then defined as the minimum time-to-collision obtained across all units.

\section{Scenario description in CARLA} \label{sec:carla}

\begin{figure*}
\centering
\begin{minipage}{0.48\textwidth}
    \centering
    \includegraphics[scale=0.2]{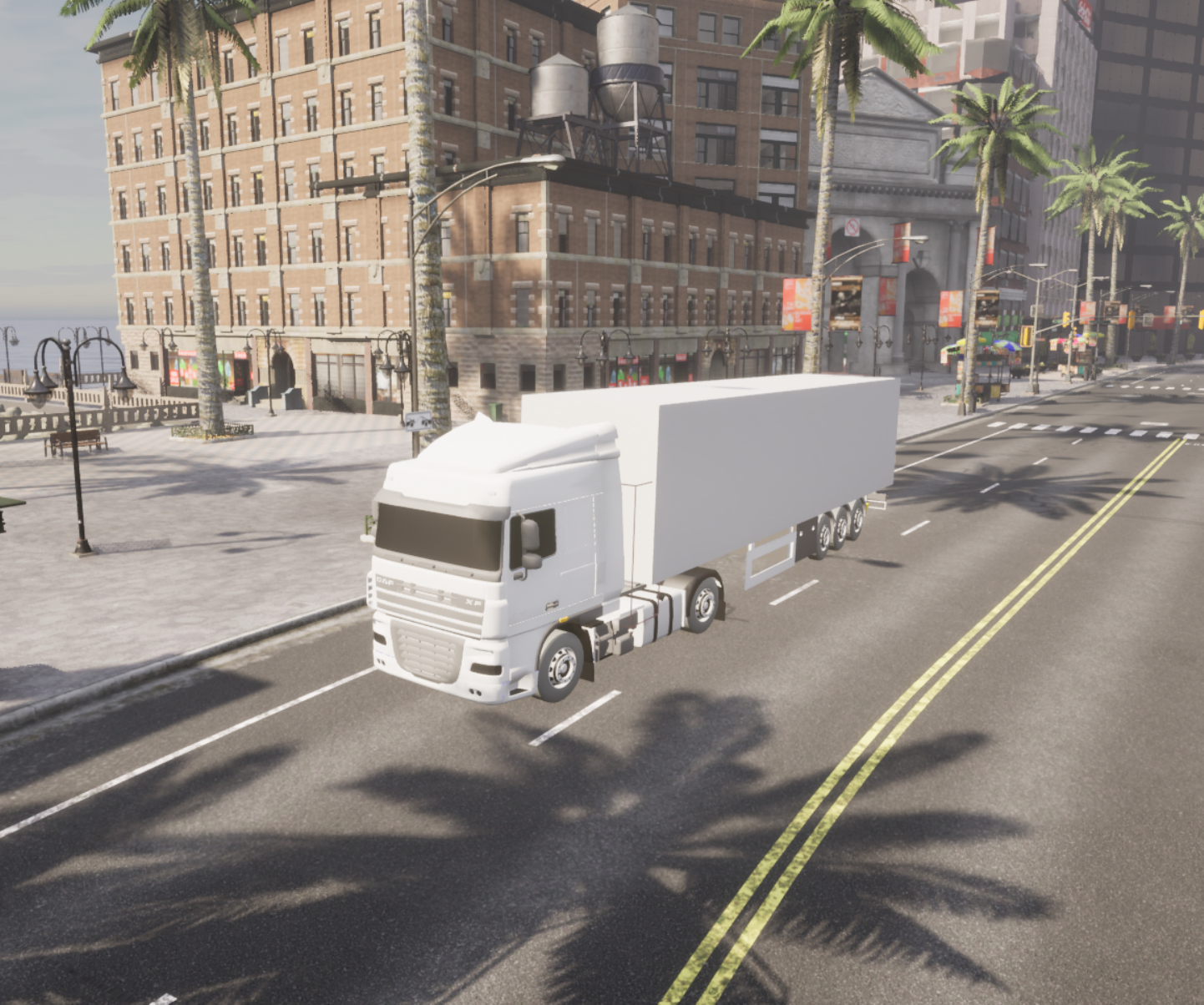}
\end{minipage}
\begin{minipage}{0.48\textwidth}
    \centering
    \textcolor{black}{\begin{tabular}{@{}cccc@{}}
    \toprule
     & \multicolumn{3}{c}{\textbf{Trailers}} \\ \cmidrule(l){2-4} 
    \textbf{Parameters} & \textbf{10 m} & \textbf{13 m} & \textbf{15 m} \\ \midrule
    $l_0$ (m)    & 6.71     & 5.85     & 6.71     \\
    $l_1$ (m)    & 10.11    & 13.18    & 14.90    \\
    $w_0$ (m)    & 3.27 & 2.91 & 3.27 \\
    $w_1$ (m)    & 2.5 & 2.51 & 2.5 \\
    $l_{fa}$ (m) & 4.4      & 3.7      & 4.4      \\
    $l_{ra}$ (m) & 1.3      & 1.65     & 1.7      \\ 
    \bottomrule
    \end{tabular}}
\end{minipage}

\caption{Tractor-semitrailer in CARLA (Left) and \textcolor{black}{dimensions of the evaluated tractor-semitrailers (Right).}}
\label{fig:carlaVM}
\end{figure*}

The proposed measures are evaluated through cut-in, curve, and roundabout scenarios implemented in the CARLA simulation environment, each leading to either a sideswipe or a rear-end collision. Each scenario involves two actors: a car and a tractor-semitrailer. While the car is available in CARLA’s standard vehicle library, a high-fidelity tractor-semitrailer model is not present in the library. To address this, a realistic tractor-semitrailer model was developed based on the work of \citet{attard2024autonomous}, and further validated and enhanced as detailed in \citet{behera2025CARLA}. This improved model is employed throughout the present analysis. \textcolor{black}{Figure~\ref{fig:carlaVM} illustrates the tractor-semitrailer and the corresponding dimensions used within the CARLA environment. The car used within CARLA has a length of $4.79$ m and a width of $2.16$ m.}

\begin{figure*}
	\centering
    \begin{minipage}{0.49\textwidth}
		\centering
		\includegraphics[scale=0.5]{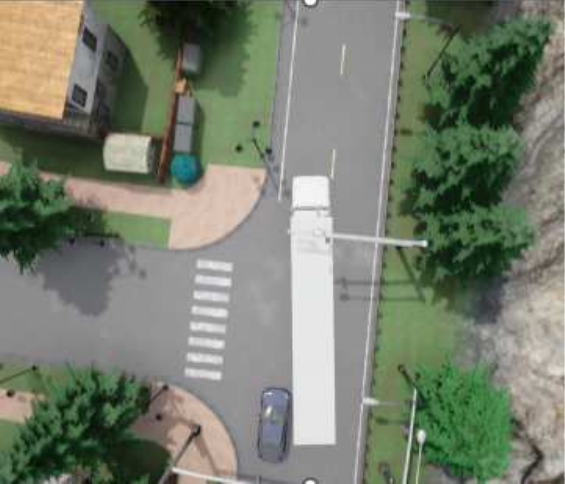}
		\subcaption{Sideswipe collision.}
		\label{fig:scollision}
	\end{minipage}
	\begin{minipage}{0.49\textwidth}
		\centering
		\includegraphics[scale=0.18]{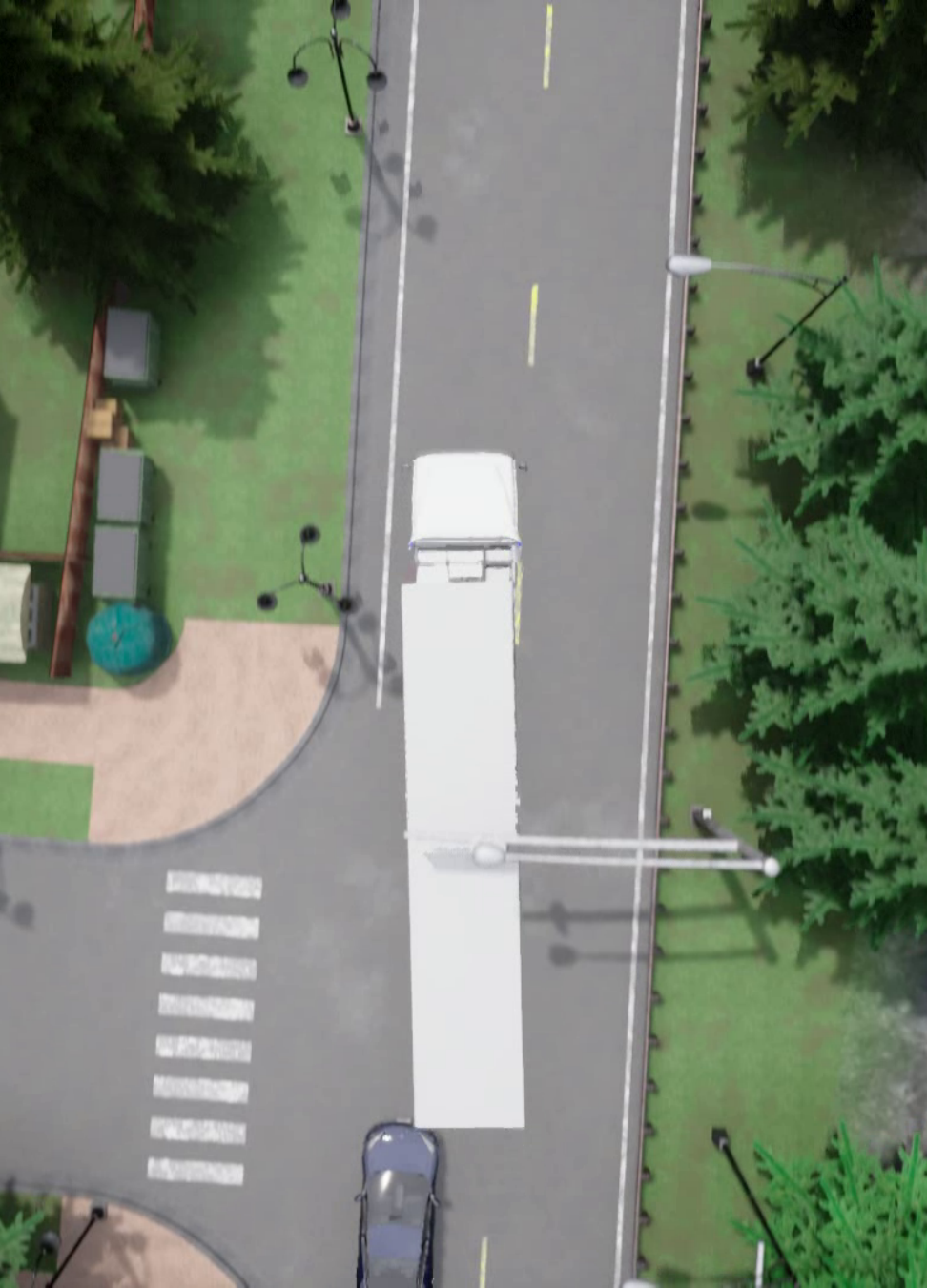}
		\subcaption{Rear-end collision.}
		\label{fig:rcollision}
	\end{minipage}
    \begin{minipage}{0.99\textwidth}
		\centering
		\includegraphics[width=1\textwidth]{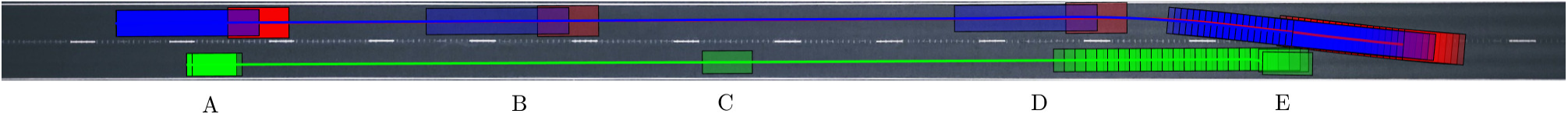}
		\subcaption{Trajectories in sideswipe collision.}
		\label{fig:sideswipe_collision}
	\end{minipage}
    	\begin{minipage}{0.99\textwidth}
		\centering
		\includegraphics[width=1\textwidth]{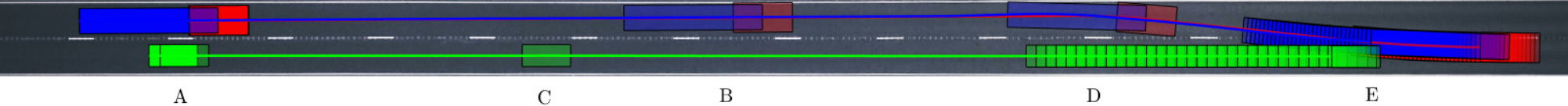}
		\subcaption{Trajectories in rear-end collision.}
		\label{fig:rearend_collision}   
	\end{minipage}
    \caption{(a, b) Two types of collisions observed in cut-in scenarios in CARLA. (c, d) Vehicle trajectories simulated in CARLA; Red: Tractor, Blue: Semitrailer, and Green: Car.}
 \label{fig:carlaVars1}
\end{figure*}

The primary objective of the evaluation is to detect sideswipe and rear-end collisions using the proposed measures, $\text{TTC}_\text{2D}^\text{AV}$ and $\text{MTTC}_\text{2D}^\text{AV}$. Accordingly, the vehicle trajectories, including speed profiles and steering angle inputs, are carefully designed to produce the intended collision outcomes within the simulation environment. 

Figures~\ref{fig:scollision} and ~\ref{fig:rcollision} show the two types of collision in cut-in scenarios. The corresponding trajectories are shown in Figures~\ref{fig:sideswipe_collision} and~\ref{fig:rearend_collision} respectively. The labeled positions indicate the vehicles' relative locations throughout the sequence. Point $A$ represents the starting point for both the car and the tractor-semitrailer. The tractor-semitrailer accelerates with a constant acceleration until it reaches a reference speed, which is set to either 60 or 80 km/h. The car has a reference speed of 80 km/h in all simulations, and a constant acceleration larger than the tractor-semitrailer's acceleration. Hence, in every scenario, the tractor-semitrailer starts its movement first. Once it reaches a specific point, depicted as point $B$, the car moves from rest at point $A$. The tractor-semitrailer initiates a cut-in maneuver at point $D$, while the car is at point $C$, and the collision occurs at point $E$. The position of point $D$ is chosen randomly, leading to a different steering angle and profile for the tractor-semitrailer in each scenario. The simulations are conducted with three different trailer lengths: $11$ m, $13$ m, and $15$ m. A total of $30$ scenarios are simulated, with $10$ scenarios for each trailer length.

 \begin{figure*}
     \centering 
	\begin{minipage}{0.45\textwidth}
		\centering
		\includegraphics[scale=0.65]{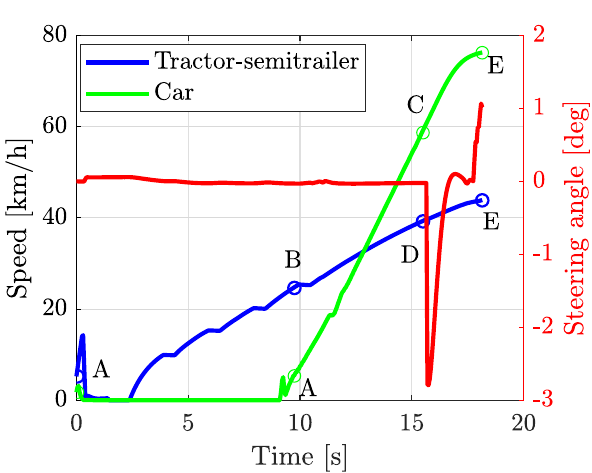}
		\subcaption{Inputs in sideswipe collision.}
		\label{fig:inputs_sideswipe}
	\end{minipage}
	\begin{minipage}{0.54\textwidth}
		\centering
		\includegraphics[scale=0.65]{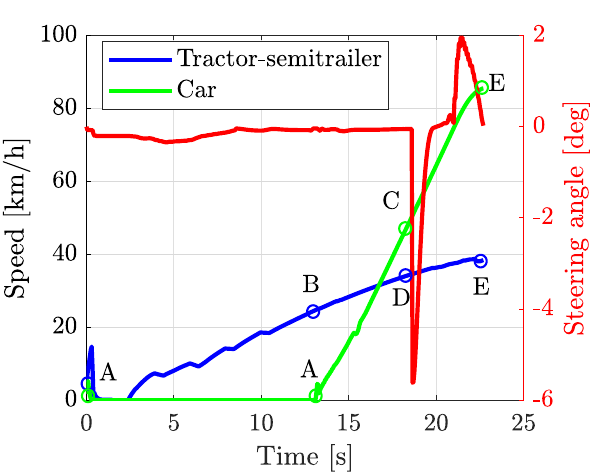}
		\subcaption{Inputs in rear-end collision.}
		\label{fig:inputs_rearend}
	\end{minipage}
	\begin{minipage}{0.45\textwidth}
		\centering
		\includegraphics[scale=0.65]{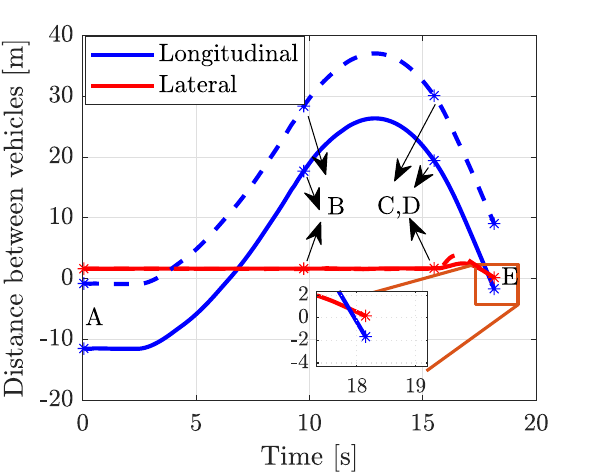}
		\subcaption{Distance in sideswipe collision.}
		\label{fig:distance_sideswipe}
	\end{minipage}
	\begin{minipage}{0.54\textwidth}
		\centering
		\includegraphics[scale=0.65]{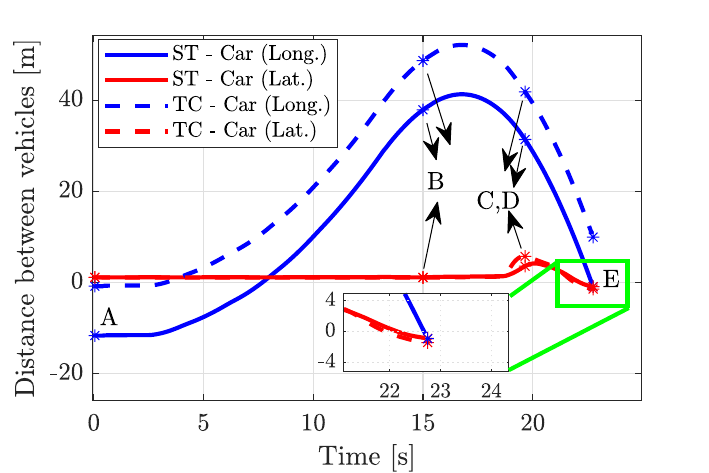}
		\subcaption{Distance in rear-end collision.}
		\label{fig:distance_rearend}
	\end{minipage}
	\caption{(a, b) Vehicle inputs in the cut-in scenario: speed of the tractor-semitrailer and the car, and the steering angle at the tire for the tractor. (c, d) Variation of the distance between the tractor (TC) and the car, and the semitrailer (ST) and the car in the scenarios.}
	\label{fig:carlaVars}
\end{figure*}

Figures~\ref{fig:inputs_sideswipe} and~\ref{fig:inputs_rearend} display the speed profiles and tire steering angles for both vehicles in a sample cut-in scenario, with key locations noted on the plot. These sideswipe and rear-end scenarios are generated using a $13$ m trailer length. \textcolor{black}{In CARLA, the acceleration of surrounding vehicles can be readily obtained through built-in functions. However, this is not straightforward in real-world scenarios. To remain consistent with real-world practices, a simple approach is followed in which acceleration is computed through successive differentiation of the speed signal. The noise in the estimated acceleration signal is subsequently smoothed using a moving average filter with a window size of $10$.} It should be noted that this introduces an approximate delay of $0.25$~s in the acceleration signal. Although a delay of this magnitude is considerable in a time-to-collision context, its influence on the present analysis is limited because the vehicles operate under constant acceleration. Nevertheless, it should be acknowledged that, in practical applications, the filtering procedure may require further refinement in order to reduce this delay. For the purposes of the conceptual evaluation presented in this work, the applied smoothing approach is adequate and does not affect the overall interpretation of the results.

Figures~\ref{fig:distance_sideswipe} and~\ref{fig:distance_rearend} present the variation of longitudinal and lateral distances between the vehicles. The longitudinal distance is defined as the separation between the rear edge of the tractor or semitrailer and the front edge of the car. A negative value indicates that the tractor or semitrailer is positioned behind the car, such as at point $A$. This distance initially increases as the car remains stationary, until around point $B$, and subsequently decreases as the car accelerates. The lateral distance represents the minimum gap between the lateral edges of the two vehicles. A negative value indicates that the vehicles are in the same lanes. Up to point $D$, the lateral distance remains approximately constant and positive, reflecting that both vehicles travel in separate lanes. The tractor-semitrailer initiates the cut-in manoeuvre at $t = 15.5$~s in the sideswipe scenario and at $t = 19$~s in the rear-end scenario. At the point of collision ($E$) in the sideswipe case, the longitudinal distance is already negative, indicating that the semitrailer is adjacent to the car. A lateral distance of zero at this point confirms a sideswipe collision. Conversely, in the rear-end collision, a negative lateral distance at point $E$ signifies that the car is positioned directly behind the semitrailer in the same lane. A longitudinal distance of zero at this point indicates a rear-end impact.

\begin{table*}
\centering
\caption{\textcolor{black}{Variables and parameters required for real-world collision prediction from a tractor-semitrailer perspective.}}
\label{tab:parameter_availability}
\begin{tabular}{ccc}
\hline
\textbf{Variables/Parameters} & \textbf{Availability} & \textbf{Typical Source} \\
\hline
Position of car & Directly measurable & Radar, LiDAR, camera, V2X \\
Speed of car & Directly measurable & Radar Doppler, object tracking, V2X \\
Acceleration of car & Estimable from speed & Numerical differentiation, Kalman filter \\
Position of tractor-semitrailer & Directly measurable & GNSS, IMU, wheel-speed sensors \\
Speed of tractor-semitrailer & Directly measurable & Wheel-speed sensors, GNSS, IMU \\
Acceleration of tractor-semitrailer & Directly measurable & IMU, vehicle CAN bus \\
Relative heading angle of car & Estimable from position & Object tracking, camera, LiDAR \\
Vehicle dimensions & Known or estimated from computer vision & Vehicle database, object classification \\
\hline
\end{tabular}
\end{table*}

\textcolor{black}{ Since the study is conducted in the CARLA simulation environment, all variables and parameters required for collision prediction are directly available through the simulator. In real-world applications, however, some quantities can be measured directly, whereas others must be estimated using onboard sensors, perception systems, and state-estimation algorithms. It may be noted that the computations are performed from the perspective of the tractor-semitrailer. The parameters and variables associated with the tractor-semitrailer can be directly measured using sensors such as an IMU or the vehicle's CAN bus. For the surrounding cars, some variables and parameters need to be estimated from cameras, radar, lidar, or V2X. Table~\ref{tab:parameter_availability} describes how each can be obtained in real-world implementations. }

\section{Discussions} \label{sec:Discussions}

This section first evaluates the performance of the proposed measures for articulated vehicles using the sample cut-in scenarios shown in Figure~\ref{fig:carlaVars1}. It then summarizes the results across all $30$ scenarios covering different trailer lengths, steering angles, and speeds. \textcolor{black}{Subsequently, the effectiveness of the proposed measures under curved trajectories is tested using two more scenarios in a roundabout and a tight turn. Finally, the computational complexity of the proposed measures is discussed at the end of this section.} It may be noted that the measure $\text{TTC}_{\text{2D-Improved}}$ is omitted from the analysis because it is intended for non-articulated vehicles, whereas this study focuses on articulated vehicles. Nonetheless, $\text{TTC}^{\text{AV}}_{\text{2D}}$, which is tailored for tractor-semitrailers, can be viewed as a natural extension of the measure $\text{TTC}_{\text{2D-Improved}}$. The validity of measure $\text{TTC}^{\text{AV}}_{\text{2D}}$ implicitly supports the efficacy of $\text{TTC}_{\text{2D-Improved}}$ for non-articulated vehicles. 

According to \cite{tidwell2015evaluation}, the perception response time of a heavy vehicle driver to a collision warning is typically about $1$ to $1.5$ s under normal conditions, with quicker responses possible when the driver is highly alert, and the warning is clear. Another study conducted by \cite{dozza2013factors} indicates that the response time ranges between $0.75$ s and $1.97$ s, with a median at $1.38$ s for daylight conditions and $1.07$ s and $2.88$ s, with a median at $1.63$ s for night conditions. Based on these studies, the analysis focuses on three time points prior to collision: $2$ s, $1.38$ s (median response time), and $1$ s. Furthermore, for visualisation purposes, all TTC values exceeding $5$ s are truncated to $5$ s, as higher values are less relevant for practical decision-making. 

\subsection{Cut-in}

\begin{figure*}
	\centering
	\begin{minipage}{0.45\textwidth}
		\centering
		\includegraphics[scale=0.65]{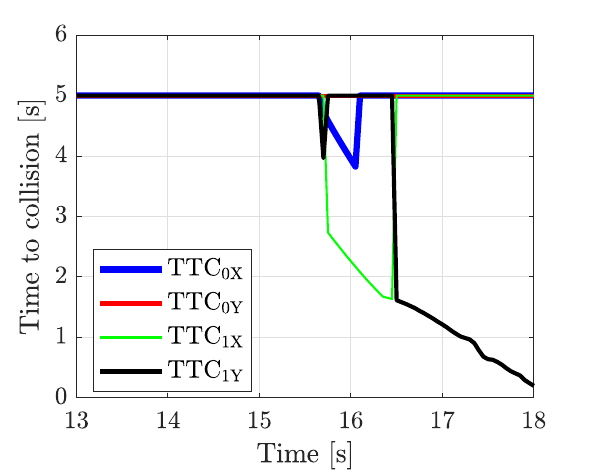}
		\subcaption{Sideswipe collision: $\text{TTC}^{\text{AV}}_{\text{2D}}$.}
		\label{fig:v2}
	\end{minipage}
	\begin{minipage}{0.45\textwidth}
		\centering
		\includegraphics[scale=0.65]{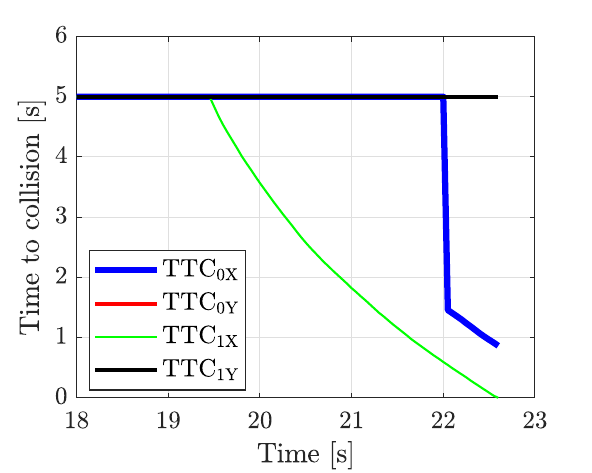}
		\subcaption{Rear-end collision: $\text{TTC}^{\text{AV}}_{\text{2D}}$.}
		\label{fig:v2_rearend}
	\end{minipage}
	\begin{minipage}{0.45\textwidth}
		\centering
		\includegraphics[scale=0.65]{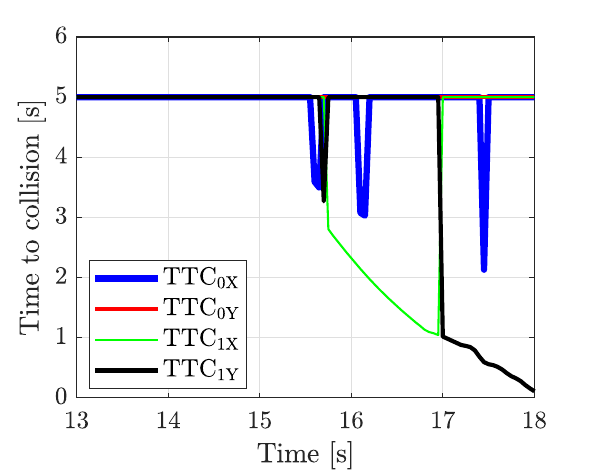}
		\subcaption{Sideswipe collision: $\text{MTTC}^{\text{AV}}_{\text{2D}}$.}
		\label{fig:v3}
	\end{minipage}
	\begin{minipage}{0.45\textwidth}
		\centering
		\includegraphics[scale=0.65]{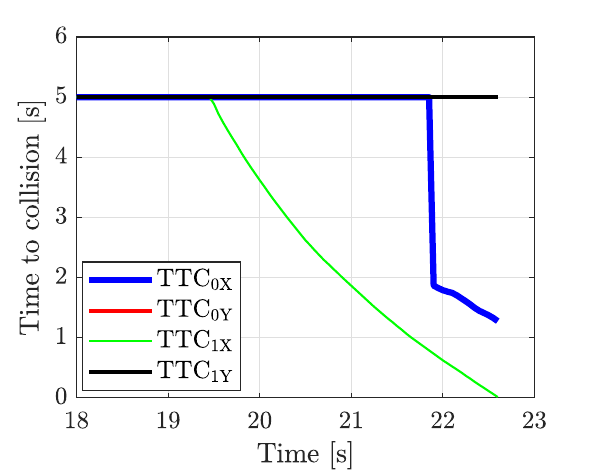}
		\subcaption{Rear-end collision: $\text{MTTC}^{\text{AV}}_{\text{2D}}$.}
		\label{fig:v3_rearend}
	\end{minipage}
	\begin{minipage}{0.45\textwidth}
		\centering
		\includegraphics[scale=0.65]{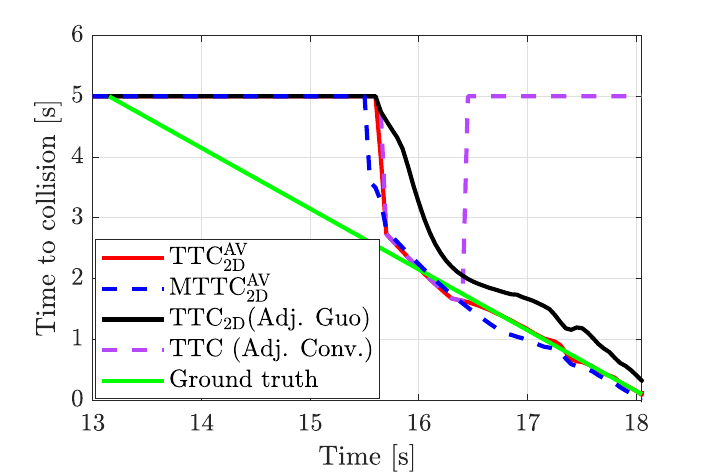}
		\subcaption{Sideswipe collision: Comparison of all measures.}
		\label{fig:ttc2d}
	\end{minipage}
	\begin{minipage}{0.45\textwidth}
		\centering
		\includegraphics[scale=0.65]{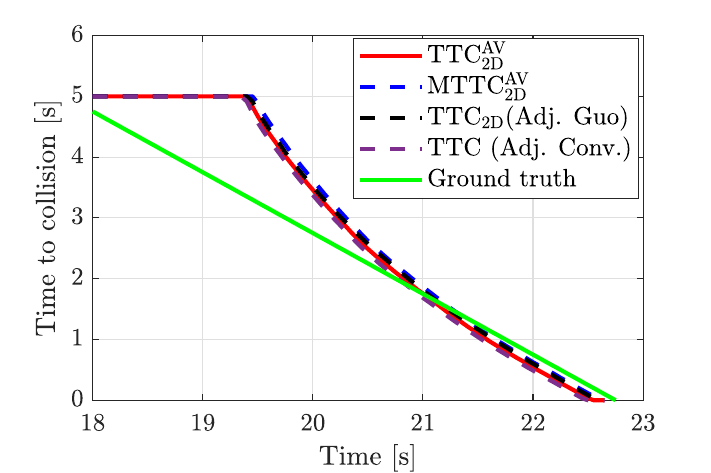}
		\subcaption{Rear-end collision: Comparison of all measures.}
		\label{fig:ttc2d_rearend}
	\end{minipage}
	\caption{Comparison of TTC in the longitudinal and lateral directions in a cut-in scenario between the car and the tractor, as well as the car and the semitrailer, for (a, b) $\text{TTC}^{\text{AV}}_{\text{2D}}$, and (c, d) $\text{MTTC}^{\text{AV}}_{\text{2D}}$. (e, f) Comparison of the proposed measures with the conventional TTC and TTC$_{\text{2D}}$ from \cite{guo2023modeling}, both adjusted for the tractor–semitrailer, along with the ground-truth TTC computed from the known trajectory data.}
	\label{fig:TTCValid}
\end{figure*}

Figure~\ref{fig:TTCValid} illustrates the performance of the proposed measures in cut-in scenarios presented in Figure~\ref{fig:carlaVars}. Figures~\ref{fig:v2} through~\ref{fig:v3_rearend} present the TTC predictions in both longitudinal and lateral directions for the car-tractor and car-semitrailer, across the two collision scenarios, for both $\text{TTC}^{\text{AV}}_{\text{2D}}$ and $\text{MTTC}^{\text{AV}}_{\text{2D}}$. In the sideswipe collision scenario, both measures accurately predict a sideswipe collision (indicated by the black solid line) between the car and the semitrailer. Notably, the initial TTC predictions predicted by these measures suggest a potential rear-end collision (green solid line) between the car and the semitrailer, which subsequently transitions to a sideswipe prediction. This shift in the predicted collision type results from the rapid acceleration of the car. The timing of this transition differs between the two measures, occurring approximately $0.5$~s later in $\text{MTTC}^{\text{AV}}_{\text{2D}}$ compared to $\text{TTC}^{\text{AV}}_{\text{2D}}$. Additionally, minor fluctuations are observed in the predictions of $\text{MTTC}^{\text{AV}}_{\text{2D}}$, likely resulting from noise in the acceleration signal. In the rear-end collision scenario, both measures predominantly identify a rear-end collision, as indicated by the green solid line (semitrailer-car) and the blue solid line (tractor-car). Among these, the TTC between the semitrailer and the car is consistently lower, signifying its relevance as the primary collision interaction in this case. 

Figures~\ref{fig:ttc2d} and~\ref{fig:ttc2d_rearend} present the computed TTC predictions obtained from the proposed measures. Each TTC value is obtained by taking the minimum of the time-to-collision predictions in the longitudinal and lateral directions. The proposed measures are compared against the conventional TTC and the existing $\text{TTC}_{\text{2D}}$ measure introduced by \citet{guo2023modeling}. Since the original $\text{TTC}_{\text{2D}}$ does not account for a tractor–semitrailer, it is adjusted here for the comparison. The adjusted implementation computes the TTC$_{\text{2D}}$ separately for the tractor and the semitrailer. The final TTC$_{\text{2D}}$ value for the articulated vehicle is then defined as the minimum of these two results. In this formulation, vehicle orientations are not taken into account. The conventional TTC is similarly adjusted to incorporate articulation effects by evaluating the longitudinal motion of both the tractor and the semitrailer. Furthermore, for reference, the ground truth TTC based on the known trajectory data is also shown in the figures.

In the sideswipe collision, the conventional TTC initially predicts a potential collision. However, after approximately $16.4$ s, it indicates no collision, implying that the vehicles would cross the same location at different times. While this correctly reflects the absence of a rear-end collision, it fails to detect the actual sideswipe event. In contrast, the proposed measures successfully identify the sideswipe collision. They give nearly identical results and demonstrate better performance compared to both the adjusted TTC and the adjusted $\text{TTC}_{\text{2D}}$. For instance, at $t = 16$~s, both $\text{TTC}^{\text{AV}}_{\text{2D}}$ and $\text{MTTC}^{\text{AV}}_{\text{2D}}$ predict a TTC of approximately $2$ s, which aligns well with the actual collision time of $18$ s (see green line). In comparison, the adjusted $\text{TTC}_{\text{2D}}$ overestimates the time-to-collision at this point, reporting a value of $3$ s. In the rear-end collision, all measures make approximately the same predictions. This convergence occurs because the TTC in the longitudinal direction is computed using comparable formulations across all measures. The predicted time-to-collision closely matches the actual collision time during the final $2$ s preceding the impact. Furthermore, the influence of the acceleration-based enhancements introduced in $\text{MTTC}^{\text{AV}}_{\text{2D}}$ is not apparent in either scenario.

\begin{figure*}
	\centering
	\begin{minipage}{0.45\textwidth}
		\centering
		\includegraphics[scale=0.65]{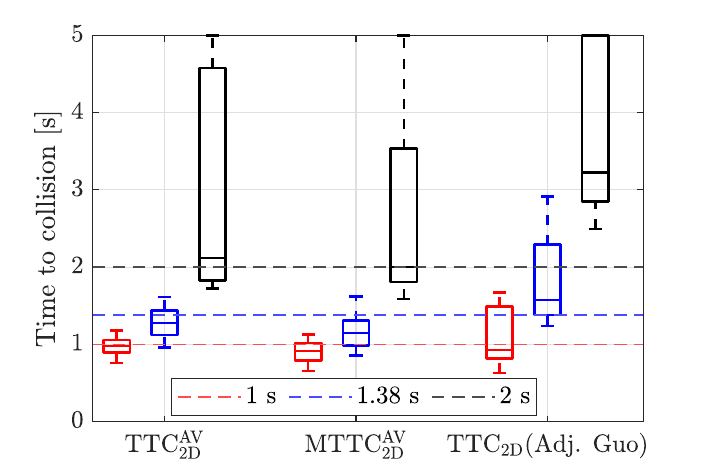}
		\subcaption{Sideswipe collisions (15 scenarios).}
		\label{fig:Lateral}
	\end{minipage}
	\begin{minipage}{0.45\textwidth}
		\centering
		\includegraphics[scale=0.65]{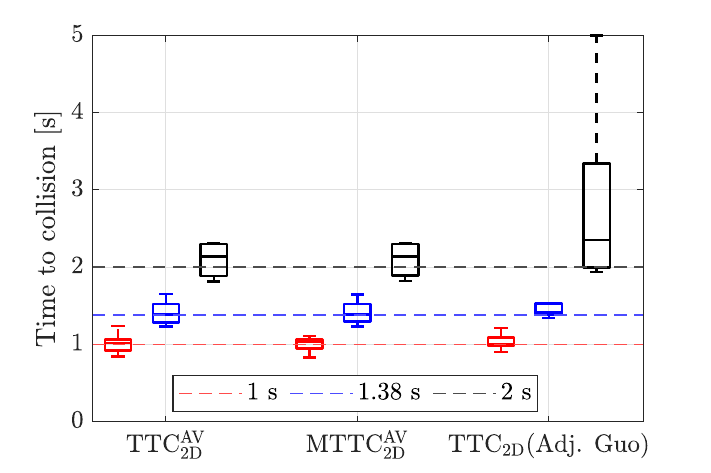}
		\subcaption{Rear-end collisions (15 scenarios).}
		\label{fig:Rear}
	\end{minipage}
	\caption{Time-to-collision predicted at $2$ s, $1.38$ s and $1$ s prior to the collision for different trailer lengths, obtained from the proposed measures and TTC$_{\text{2D}}$ (Adj. Guo). The dashed lines indicate the ground truth.}
	\label{fig:LateralRear}
\end{figure*}

\begin{figure*}
	\centering
	\begin{minipage}{0.45\textwidth}
		\centering
		\includegraphics[scale=0.65]{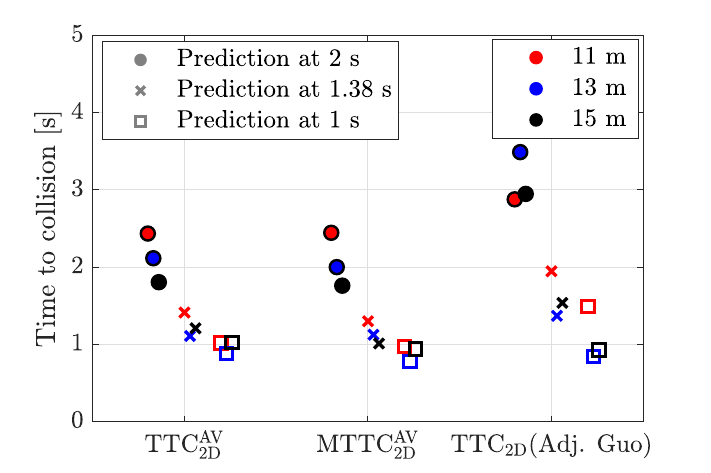}
		\subcaption{Sideswipe collisions.}
		\label{fig:Lateral_111315}
	\end{minipage}
	\begin{minipage}{0.45\textwidth}
		\centering
		\includegraphics[scale=0.65]{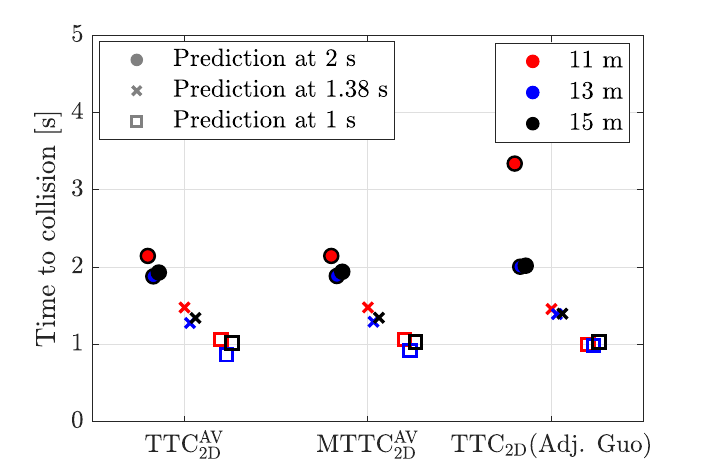}
		\subcaption{Rear-end collisions.}
		\label{fig:Rear_111315}
	\end{minipage}
	\caption{Median of predictions for different trailer lengths at $2$ s, $1.38$ s and $1$ s prior to the collision. Each data point for a given trailer length represents five scenarios per collision type.}
	\label{fig:LateralRear_111315}
\end{figure*}

Figure \ref{fig:LateralRear} shows the analysis for all the trailer lengths across both collision types, and for all the 30 scenarios. The predictions are reported at $2$ s, $1.38$ s (median response time obtained from \cite{dozza2013factors}), and $1$ s prior to the collision obtained using the proposed measures and the existing formulation (after adjusting for tractor-semitrailer) from \citet{guo2023modeling}. All formulations predicted that a collision would occur. However, for sideswipe collision scenarios, the existing formulation only identified $7$ of the $15$  collisions as the correct type of collision, i.e., sideswipe. In comparison, the proposed measures correctly classified $14$ of $15$ sideswipe collisions, with one misclassified as a rear-end collision. This outlier involved a late lane change by the tractor–semitrailer. This is also the only case that resulted in a predicted TTC of $5$ s at $2$ s before collision in Figure \ref{fig:Lateral}. Regardless of the misclassifications, all the scenarios are included in Figure \ref{fig:Lateral} since they predicted the collision. For rear-end collision scenarios shown in Figure \ref{fig:Rear}, all formulations predicted the collision type correctly. \textcolor{black}{Tables \ref{tab:lateral_ttc} and \ref{tab:rear_ttc} present the median predictions and interquartile (IQR) regions  for all trailer lengths and all 30 scenarios across both collision types. The interquartile region, represented by the width of the box in the box plots, indicates the consistency of the predicted TTC values.}

\begin{table*}
\centering
\caption{\textcolor{black}{Median TTC predictions with absolute percentage error relative to the ground truth and corresponding IQR (interquartile range) for sideswipe collisions.}}
\label{tab:lateral_ttc}
\setlength{\tabcolsep}{4pt}
\begin{tabular}{ccccccccc}
\hline
\multirow{2}{*}{Ground Truth (s)} &
\multirow{2}{*}{Trailer Length (m)} &
\multicolumn{2}{c}{TTC$^{\text{AV}}_{\mathrm{2D}}$} &
\multicolumn{2}{c}{MTTC$^{\text{AV}}_{\mathrm{2D}}$} &
\multicolumn{2}{c}{TTC$_{\mathrm{2D}}\:\mathrm{(Adj.\: Guo)}$} \\
\cline{3-8}
& & Median (\% Error) & IQR & Median (\% Error) & IQR & Median (\% Error) & IQR \\
\hline
1.00 & 11 & 1.02 (1.5\%) & 0.116 & 0.97 (3.0\%) & 0.130 & 1.49 (49.0\%) & 1.181 \\
1.00 & 13 & 0.88 (12.0\%) & 0.129 & 0.78 (22.5\%) & 0.088 & 0.84 (15.6\%) & 0.211 \\
1.00 & 15 & 1.03 (2.5\%) & 0.131 & 0.94 (6.0\%) & 0.204 & 0.92 (7.6\%) & 0.980 \\
\hline
1.38 & 11 & 1.41 (2.2\%) & 0.275 & 1.30 (5.9\%) & 0.303 & 1.95 (41.0\%) & 1.663 \\
1.38 & 13 & 1.11 (19.8\%) & 0.265 & 1.12 (18.5\%) & 0.279 & 1.37 (0.7\%) & 0.737 \\
1.38 & 15 & 1.21 (12.5\%) & 0.309 & 1.01 (26.7\%) & 0.429 & 1.54 (11.3\%) & 0.730 \\
\hline
2.00 & 11 & 2.43 (21.7\%) & 1.809 & 2.44 (22.2\%) & 1.803 & 2.88 (43.9\%) & 2.135 \\
2.00 & 13 & 2.11 (5.7\%) & 3.061 & 2.00 (0.0\%) & 2.027 & 3.49 (74.5\%) & 1.940 \\
2.00 & 15 & 1.80 (9.8\%) & 1.329 & 1.76 (12.0\%) & 1.199 & 2.95 (47.4\%) & 1.812 \\
\hline
\end{tabular}
\end{table*}

\begin{table*}
\centering
\caption{\textcolor{black}{Median TTC predictions with percentage error relative to the ground truth and corresponding IQR (interquartile range) for rear-end collisions.}}
\label{tab:rear_ttc}
\setlength{\tabcolsep}{4pt}
\begin{tabular}{ccccccccc}
\hline
\multirow{2}{*}{Ground Truth (s)} &
\multirow{2}{*}{Trailer Length (m)} &
\multicolumn{2}{c}{TTC$^{\text{AV}}_{\mathrm{2D}}$} &
\multicolumn{2}{c}{MTTC$^{\text{AV}}_{\mathrm{2D}}$} &
\multicolumn{2}{c}{TTC$_{\mathrm{2D}}\:\mathrm{(Adj.\:Guo)}$} \\
\cline{3-8}
& & Median (\% Error) & IQR & Median (\% Error) & IQR & Median (\% Error) & IQR \\
\hline

1.00 & 11 & 1.065 (6.5\%) & 0.116 & 1.065 (6.5\%) & 0.125 & 0.95 (5\%) & 0.109 \\
1.00 & 13 & 0.865 (13.5\%) & 0.103 & 0.920 (8.0\%) & 0.138 & 0.987 (1.3\%) & 0.073 \\
1.00 & 15 & 1.015 (1.5\%) & 0.109 & 1.030 (3.0\%) & 0.104 & 1.030 (3.0\%) & 0.138 \\

\hline

1.38 & 11 & 1.478 (7.1\%) & 0.201 & 1.478 (7.1\%) & 0.206 & 1.488 (8\%) & 0.190 \\
1.38 & 13 & 1.276 (7.5\%) & 0.200 & 1.292 (6.4\%) & 0.188 & 1.391 (0.8\%) & 0.184 \\
1.38 & 15 & 1.340 (2.9\%) & 0.208 & 1.345 (2.5\%) & 0.221 & 1.397 (1.2\%) & 0.213 \\

\hline

2.00 & 11 & 2.145 (7.3\%) & 0.550 & 2.145 (7.3\%) & 0.544 & 2.341 (15.1\%) & 1.691 \\
2.00 & 13 & 1.880 (6.0\%) & 0.674 & 1.885 (5.8\%) & 0.671 & 2.15 (7.5\%) & 0.646 \\
2.00 & 15 & 1.930 (3.5\%) & 0.608 & 1.940 (3.0\%) & 0.600 & 2.019 (1.0\%) & 0.627 \\

\hline
\end{tabular}
\end{table*}

\begin{figure*}
	\centering
	\begin{minipage}{0.99\textwidth}
		\centering
		\includegraphics[scale=0.65]{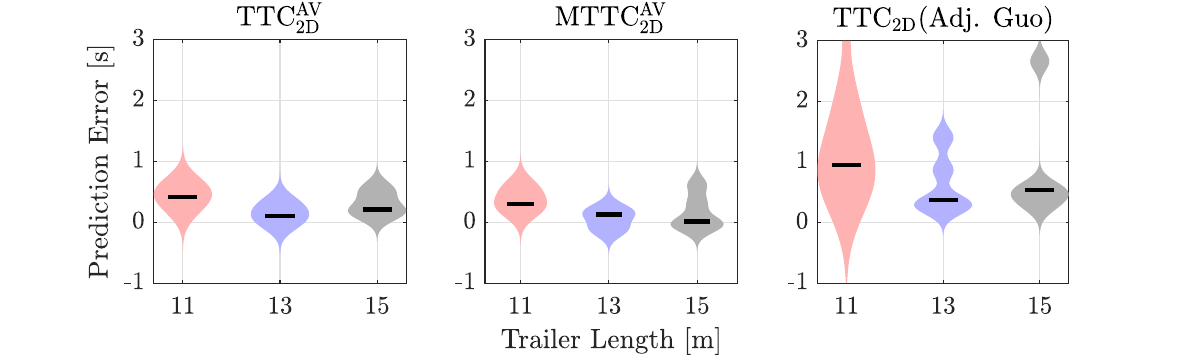}
		\subcaption{Sideswipe collisions.}
        \vspace{1mm}
		\label{fig:LateralError_138}
	\end{minipage}
	\begin{minipage}{0.99\textwidth}
		\centering
		\includegraphics[scale=0.65]{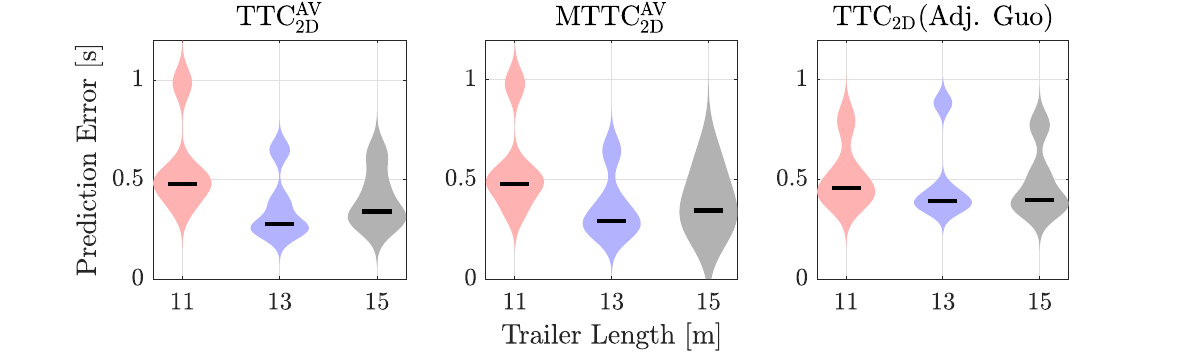}
		\subcaption{Rear-end collisions.}
		\label{fig:RearError_138}
	\end{minipage}
	\caption{\textcolor{black}{Distribution of prediction errors at $1.38$ s across varying trailer lengths. The solid line is the median. Positive prediction error indicates the predicted value is less than the ground truth and vice versa for negative error. }}
	\label{fig:Error_138}
\end{figure*}

\textcolor{black}{ For sideswipe collisions, the measure $\text{MTTC}^{\text{AV}}_{\text{2D}}$ provides relatively consistent predictions at $2$ s prior to collision, as indicated by its lower IQR in Table \ref{tab:lateral_ttc} and Figure \ref{fig:Lateral}. However, at shorter time horizons, the measure $\text{MTTC}^{\text{AV}}_{\text{2D}}$ tends to underestimate the collision time, possibly due to the constant acceleration assumption, which may exaggerate speed variations. In contrast, the measure $\text{TTC}^{\text{AV}}_{\text{2D}}$ generally provides more consistent predictions (lower IQR) with better accuracy at $1$ s and $1.38$ s prior to collision compared to the measure $\text{MTTC}^{\text{AV}}_{\text{2D}}$. It may be noted that the existing measure, TTC$_{\mathrm{2D}}\:\mathrm{(Adj.\:Guo)}$, shows lower median error percentages in some cases; however, its higher IQRs indicate greater variability. Thus, the lower median errors do not necessarily imply better performance, as the proposed measures provide more consistent predictions. For rear-end collisions, both proposed measures have similar performance (in some cases slightly worse) when compared to the existing measures. The difference between the two proposed measures is also small for both collision types, likely due to the short duration of lane changes in the cut-in scenario. This is evident from Table \ref{tab:rear_ttc} and Figures \ref{fig:Rear} and \ref{fig:Rear_111315}.}

Furthermore, as evident from the tables and figures, the IQR from all three formulations decreases as the time to collision approaches in both collision types. This behavior is expected, as vehicle trajectories become more constrained and collision geometry more deterministic over shorter horizons. This trend is also observed in Figure~\ref{fig:LateralRear_111315}, where the median predicted TTC$_{\text{2D}}$ is shown for each trailer length. The square markers (predictions at $1$ s) are generally more tightly clustered than the crosses and circles (predictions at $2$ s and $1.38$ s, respectively). Additionally, predictions for collisions involving the semitrailer of length $11$ m appear to be less accurate across all time horizons. This observation is further supported by Figure~\ref{fig:Error_138}, which shows the distribution of prediction errors, defined as the difference between ground truth and predicted values at $1.38$ s, for both sideswipe and rear-end collisions across varying trailer lengths. It can be inferred from Figure~\ref{fig:Error_138} that both proposed measures have lower prediction errors than the adjusted $\text{TTC}_{\text{2D}}$ for sideswipe collisions. Among them, $\text{MTTC}^{\text{2D}}_{\text{AV}}$ performs slightly better than $\text{TTC}^{\text{2D}}_{\text{AV}}$. All measures show similar performance for rear-end collisions. \textcolor{black}{Table \ref{tab:average_error} shows the mean absolute percentage prediction error across all $30$ scenarios computed at $1.38$ s before collision. The results indicate that the proposed measures achieve comparable prediction accuracy to the existing formulation for rear-end collisions, while providing significantly improved accuracy for sideswipe collisions, with approximately a $20$\% reduction in prediction error. }

\begin{table}
\centering
\caption{\textcolor{black}{Mean percentage prediction error of TTC predictions for sideswipe and rear-end collisions at $1.38$ s before collision.}}
\label{tab:average_error}
\begin{tabular}{lccc}
\hline
\multirow{2}{*}{Collision Type} &
\multicolumn{3}{c}{Prediction Error (\%)} \\
\cline{2-4}
& TTC$^{\text{AV}}_{\mathrm{2D}}$ & MTTC$^{\text{AV}}_{\mathrm{2D}}$ & TTC$_{\mathrm{2D}}$ (Adj. Guo) \\
\hline
Sideswipe & 9.74 & 12.18 & 32.33 \\
Rear-end  & 6.20 & 5.51 & 4.77 \\
\hline
\end{tabular}
\end{table}

\begin{figure*}
	\centering
	\includegraphics[scale=0.72]{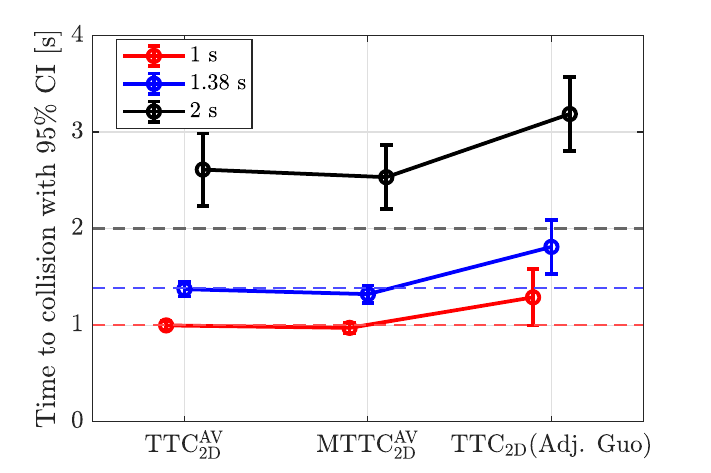}
	\caption{Error statistics of mean predictions for all $30$ scenarios obtained using bootstrap resampling. The dashed lines indicate the ground truth.}
	\label{fig:bootstrapResampling}
\end{figure*}

To quantify the statistical reliability of the proposed measures, non-parametric $95$\% confidence intervals (CIs) are obtained via $10000$-sample bootstrap resampling. Bootstrapping is chosen as the underlying error distribution is unknown, which is typical in surrogate safety evaluation, where TTC errors may exhibit skewness, multi-modality, or heavy tails. For each measure, the mean prediction is repeatedly recomputed on bootstrap resamples of the simulation data. The $2.5$ and $97.5$ percentiles of the resulting bootstrap distribution define the $95$\% confidence interval.

The resulting CIs in Figure \ref{fig:bootstrapResampling} quantify the uncertainty associated with the mean prediction of each measure. Narrow intervals indicate consistent performance across scenarios, whereas wide intervals reflect higher variability and reduced reliability. Both $\text{TTC}_\text{2D}^\text{AV}$ and $\text{MTTC}_\text{2D}^\text{AV}$ have narrow intervals and negligible biases, in contrast to $\text{TTC}_\text{2D}$ (Adj. Guo), which shows substantially wider intervals and larger bias at prediction times of $1$ s and $1.38$ s. At $2$ s, the interval widths become comparable across all the measures, although the proposed measures maintain notably lower biases. Between them, $\text{MTTC}_\text{2D}^\text{AV}$ performs slightly better at longer time horizons. 

\textcolor{black}{\subsection{Tight turn and Roundabout}}

\begin{figure*}
	\centering
    \begin{minipage}{0.49\textwidth}
		\centering
		\includegraphics[scale=0.125]{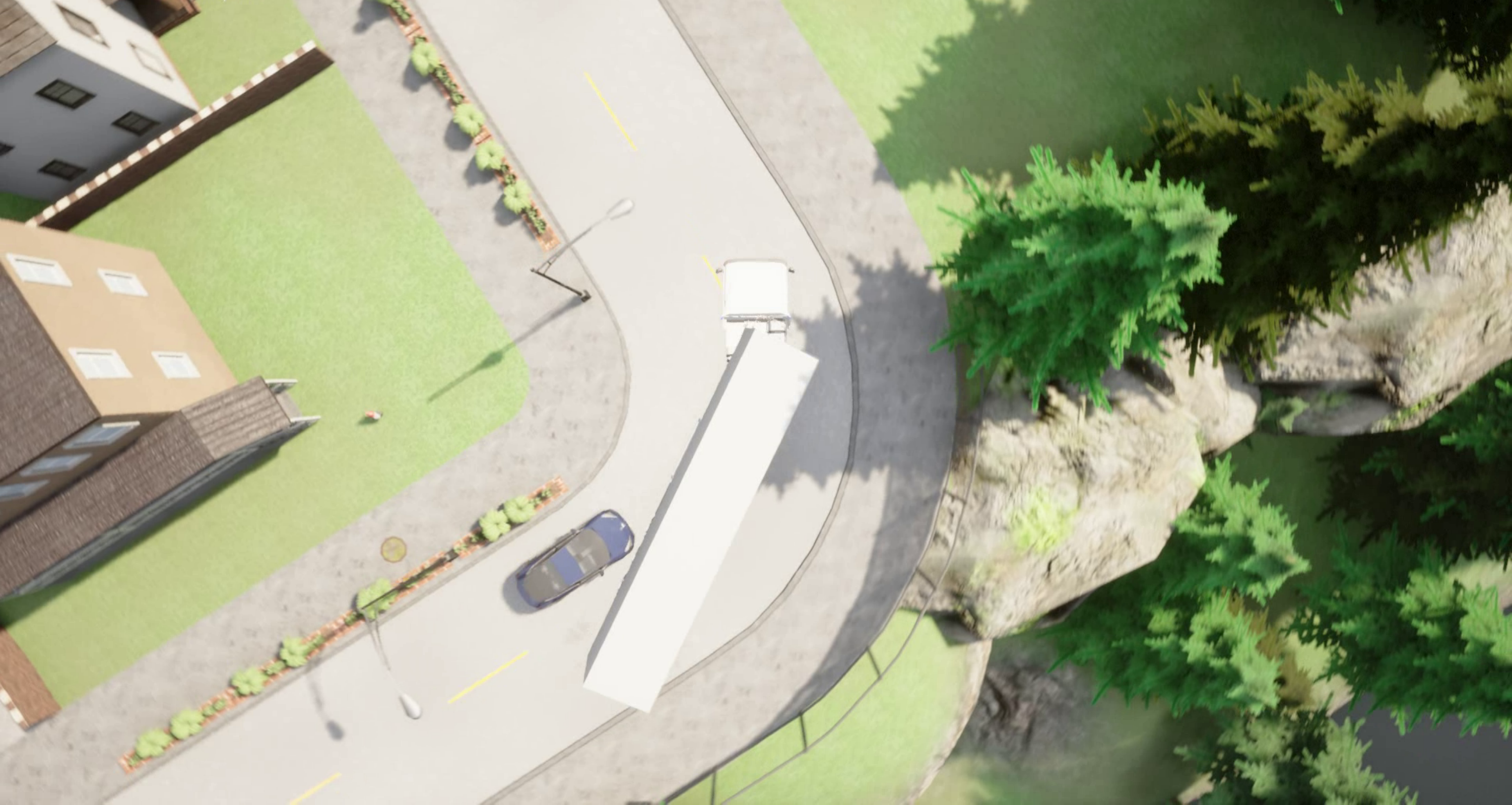}
		\subcaption{Tight turn: Sideswipe collision.}
		\label{fig:carlaCurves}
	\end{minipage}
	\begin{minipage}{0.49\textwidth}
		\centering
		\includegraphics[scale=0.125]{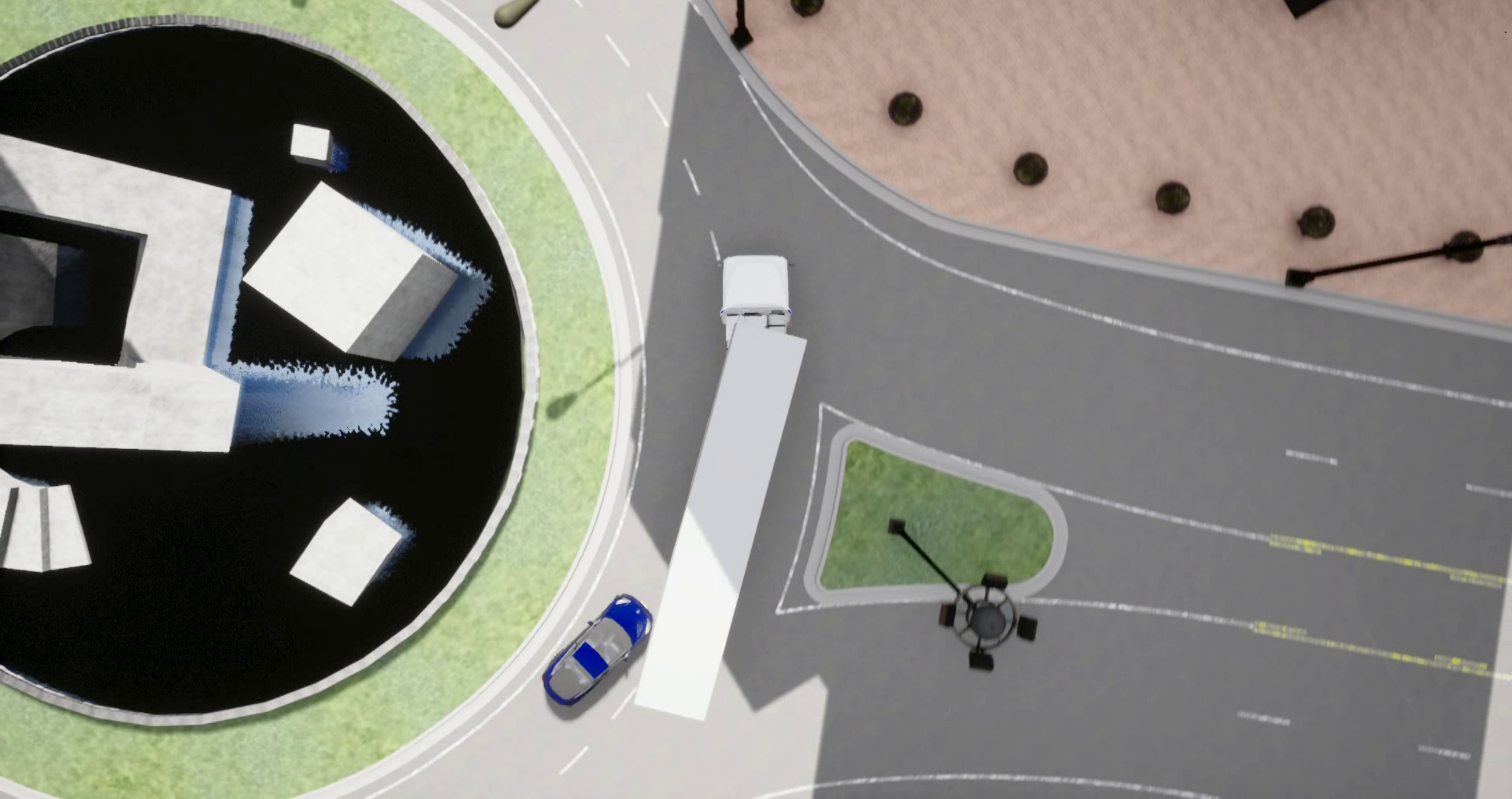}
		\subcaption{Roundabout: Sideswipe collision.}
		\label{fig:carlaRoundabout}
	\end{minipage}
	\begin{minipage}{0.49\textwidth}
		\centering
		\includegraphics[width=1\textwidth]{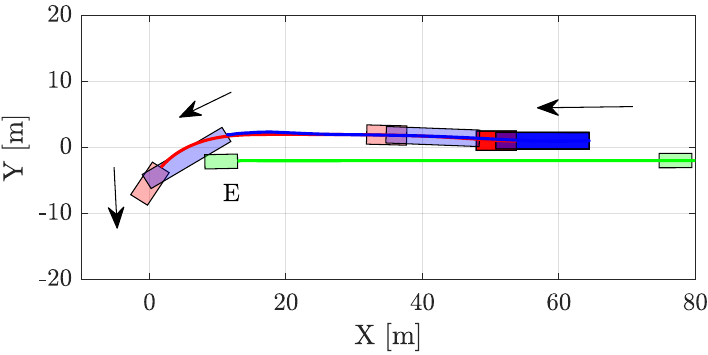}
		\subcaption{Trajectories in the tight turn.}
		\label{fig:curve}
	\end{minipage}
	\begin{minipage}{0.49\textwidth}
		\centering
		\includegraphics[width=1\textwidth]{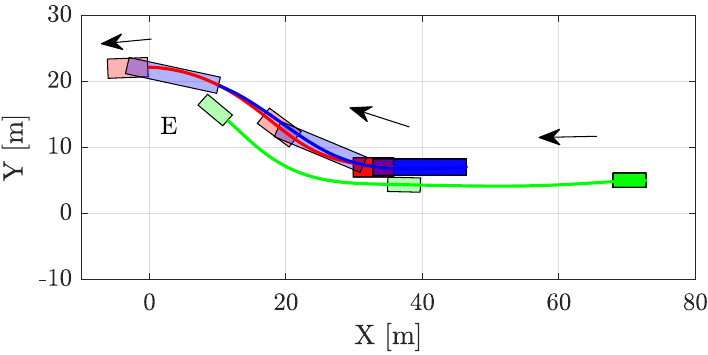}
		\subcaption{Trajectories in the roundabout.}
		\label{fig:roundabout}
	\end{minipage}
	\caption{\textcolor{black}{Scenarios with curved trajectories, used for the analysis. Red: Tractor, Blue: Semitrailer, and Green: Car.}}
	\label{fig:carlaCurveRoundabout}
\end{figure*}

\begin{figure*}
	\centering
	\begin{minipage}{0.45\textwidth}
		\centering
		\includegraphics[scale=0.68]{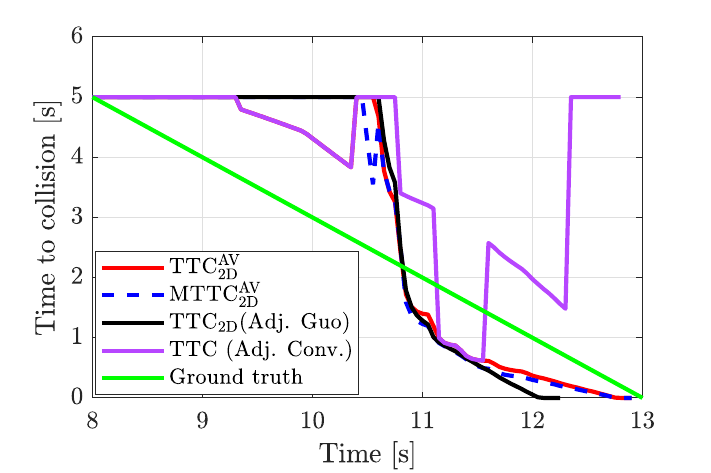}
		\subcaption{Tight turn: Comparison of all measures.}
		\label{fig:ttc2d_curve}
	\end{minipage}
	\begin{minipage}{0.45\textwidth}
		\centering
		\includegraphics[scale=0.68]{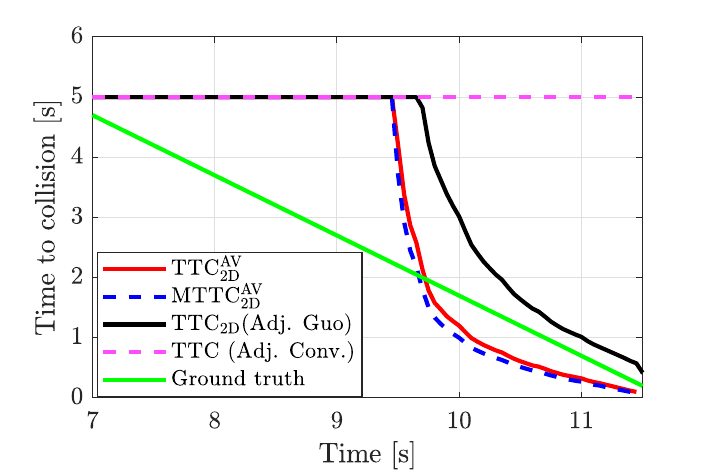}
		\subcaption{Roundabout: Comparison of all measures.}
		\label{fig:ttc2d_roundabout}
	\end{minipage}
	\caption{\textcolor{black}{ Comparison of the proposed measures with the conventional TTC and TTC$_{\text{2D}}$ from \cite{guo2023modeling}, both adjusted for the tractor–semitrailer, along with the ground-truth TTC computed from the known trajectory data.}}
	\label{fig:TTCValidCR}
\end{figure*}

\textcolor{black}{To demonstrate that the proposed measures are also effective under curved trajectories, a brief analysis is conducted using two additional scenarios in a roundabout and a tight turn. In these cases, sideswipe collisions are commonly observed as the semitrailer encroaches into the adjacent lane due to offtracking. Figures~\ref{fig:carlaCurves} and~\ref{fig:carlaRoundabout} illustrate sideswipe collisions in these scenarios. The corresponding vehicle trajectories are shown in Figures~\ref{fig:curve} and~\ref{fig:roundabout}, respectively. Point $E$ denotes the location at which the collision occurs. As evident from the figures, a part of the semitrailer occupies the left lane in both cases, obstructing the car’s path and ultimately leading to a sideswipe collision.}

\textcolor{black}{Figure~\ref{fig:TTCValidCR} illustrates the performance of the proposed measures in the simulated roundabout and tight turn. Both measures predict the collision type, i.e., sideswipe collision, correctly in both scenarios. The proposed measures perform much better compared to the conventional TTC in both scenarios, see Figure~\ref{fig:ttc2d_curve}-\ref{fig:ttc2d_roundabout}. The conventional TTC fails to predict any collision in the roundabout scenario, whereas the existing $\text{TTC}_{\text{2D}}$ provides predictions similar to the proposed measures. However, it overestimates the collision time in the roundabout scenario. On the contrary, the proposed measures underestimate it relative to the ground truth. While a discrepancy between the proposed measures and the ground truth is also observed, it takes the form of a slight underestimation. This behavior is preferable to overestimation, as it is more conservative and better accounts for delayed driver response. For the tight turn scenario, both the existing and proposed measures underestimate the collision time. There is not a significant difference between $\text{TTC}_\text{2D}^\text{AV}$ and $\text{MTTC}_\text{2D}^\text{AV}$, likely due to a less aggressive acceleration profile.}  

\textcolor{black}{Similar to the roundabout and tight turn, the proposed measures can also be applied to crossing conflicts at intersections where vehicles approach the conflict point with different headings. The rotation matrix defined in (\ref{eq:r1comb}) adjusts the projected vehicle dimensions as well as the longitudinal and lateral velocity components to account for differences in heading between interacting vehicles. Consequently, the formulation is not restricted to analysed scenarios and can be used to identify potential collisions arising from intersecting trajectories. However, the conventional classification of collision types as rear-end or sideswipe no longer applies in such cases. For example, if the front of the car impacts the side of the tractor or semitrailer at an intersection (the heading difference is around $90$ deg), the resulting conflict can be classified as a side-impact collision (\cite{kumaresan2006biomechanics}). Nevertheless, the proposed measures can still be used to estimate the time-to-collision in such scenarios, as differences in vehicle heading are explicitly considered through the coordinate transformation. }

\textcolor{black}{\subsection{Computation time}}

\textcolor{black}{An important aspect of the proposed measures is computation time. Although the derived mathematical formulations are effective, real-world implementation requires continuous high-frequency updates for collision avoidance systems to remain reliable. For the proposed measures, the total computation time comprises the time required to evaluate the collision time for both the tractor–car and semitrailer-car interactions. The tractor-car interaction is governed by a small set of analytical expressions, making it computationally efficient. Similar formulations are also used for the semitrailer-car interaction; however, a forward Euler discretization method is employed to propagate the dynamics, which cumulatively incurs a slightly higher computational cost. Nevertheless, the total computation time for both measures remains on the order of $10^{-5}$ s for one prediction loop. This is in line with the requirements of many real-time systems, as such systems are expected to operate at a frequency of $10$ Hz or higher \citep{ISO15623}.}

\section{Conclusions and Future work} \label{sec:conclusions}
This paper presents the development of two-dimensional time-to-collision measures for articulated vehicles, aimed at predicting both sideswipe and rear-end collisions. First, the existing TTC$_{\text{2D}}$ measure for non-articulated vehicles is improved to include vehicle orientation. Subsequently, two new measures are proposed, $\text{TTC}_\text{2D}^\text{AV}$ and $\text{MTTC}_\text{2D}^\text{AV}$, that can be used for articulated vehicles. The former introduces articulation constraints while assuming constant speed and heading. The latter replaces the constant-speed assumption with constant acceleration, forming the two-dimensional analogue of the modified TTC. The performance of these measures for articulated vehicles is then evaluated using cut-in, curve, and roundabout scenarios created in CARLA.

The proposed measures demonstrate several clear advantages over existing formulations, more specifically, $\text{TTC}_{\text{2D}}$ developed by \cite{guo2023modeling} and conventional TTC. The proposed measures provide a framework for incorporating articulated vehicles, which is not addressed in existing TTC-based literature. Unlike conventional TTC, which fails to detect sideswipe collisions, the proposed measures accurately identify both the type and timing of such collisions. Across $30$ simulated cut-in scenarios, they correctly classify $14$ out of $15$ sideswipe collisions, compared to only $7$ out of $15$ for the existing $\text{TTC}_{\text{2D}}$ formulation. The proposed measures have similar performance to the existing measure in detecting rear-end collisions. However, it demonstrates a substantial improvement in detecting lateral collision risks. \textcolor{black}{The proposed measures achieve an average of $20\%$ reduction in prediction error compared with the existing formulation for sideswipe collisions.} In general, for sideswipe collisions, the collision time predicted by the proposed measures aligns more closely with the ground truth compared to existing methods, which tend to overestimate collision times. The advantages are further confirmed in more complex scenarios, such as tight turns and roundabouts. The proposed measures correctly predict the collision type, whereas the conventional TTC formulation fails to detect collisions in both the roundabout and the tight turn. Overall, the benefits of the acceleration-based enhancement in $\text{MTTC}_\text{2D}^\text{AV}$ can be observed at longer time-to-collision horizons where acceleration has more influence. However, since most of the simulated scenarios are short-duration manoeuvres and do not have an aggressive acceleration profile, the differences between the two proposed measures remain relatively small. 

\textcolor{black}{Although the proposed measures $\text{TTC}_\text{2D}^\text{AV}$ and $\text{MTTC}_\text{2D}^\text{AV}$ are developed and evaluated from a collision prediction perspective, their potential applications extend to several transportation domains. In freight operations, the measures can support safer manoeuvre planning for articulated heavy vehicles during lane changes, merging, intersection negotiation, and operation in constrained environments where off-tracking and trailer swing may create lateral collision risks.  The measures may also be incorporated into microscopic traffic simulation to evaluate the safety impacts of increasing long combination vehicles without relying solely on historical crash data. From a human factors perspective, the ability to distinguish between rear-end and sideswipe conflicts may facilitate the development of more intuitive driver warning and assistance systems that better reflect the nature of the underlying risk. Nevertheless, practical deployment requires reliable estimation of vehicle states, articulation angles, and surrounding traffic trajectories under real-world sensing uncertainties, which remains an important area for future research.}

\textcolor{black}{Even though the current study relies on simulation data, the estimation errors were inevitable and handled using a simple filter. Future work should therefore include sensitivity analyses to evaluate the robustness of $\text{MTTC}_\text{2D}^\text{AV}$ to estimation errors, particularly in acceleration components. This would help determine the extent to which the proposed measure can tolerate such uncertainties before its predictive accuracy degrades.} Future research can further extend the proposed measures to relax the constant heading assumption in the prediction horizon. Subsequently, for example, a constant curvature model can be used for the heading of the tractor. Another research direction includes using the measure in the closed-loop simulation. For example, one can use it as a constraint to the MPC controller for a motion planning problem where TTC is mostly used. It is also interesting to deploy this measure to real-world vehicles and measure its efficacy. 

\printcredits

\section*{Data Availability}
The code and simulation data supporting this study are publicly available at \cite{behera2025simulation}.

\section*{Declaration of competing interest}
The authors declare that they have no known competing financial interests or personal relationships that could have appeared to influence the work reported in this paper.

    \section*{Funding}
Co-funded by the European Union. Views and opinions expressed are however those of the author(s) only and do not necessarily reflect those of the European Union or European Climate, Infrastructure and Environment Executive Agency (CINEA). Neither the European Union nor the granting authority can be held responsible for them. European project ROADVIEW, grant no. 101069576.

\appendix
\renewcommand{\theequation}{\Alph{section}.\arabic{equation}}
\setcounter{equation}{0}
\section{Appendix: Yaw angle computation} \label{app:Method2}

A vehicle is a system with non-holonomic constraints \citep{minaker2010automatic}. Let $\psi$ be the vehicle's orientation, and $X$ and $Y$ be the position coordinates of the axle in the global reference system. The  non-holonomic constraint corresponding to one of the axles of the vehicle can be written as:
\begin{equation}
	\dot{X} \sin \psi - \dot{Y} \cos \psi = 0.
	\label{eq: nonhol_const}
\end{equation}
Similar expressions can be obtained for a tractor-semitrailer combination. 

\begin{figure}
	\centering
	\includegraphics[width=0.45\textwidth]{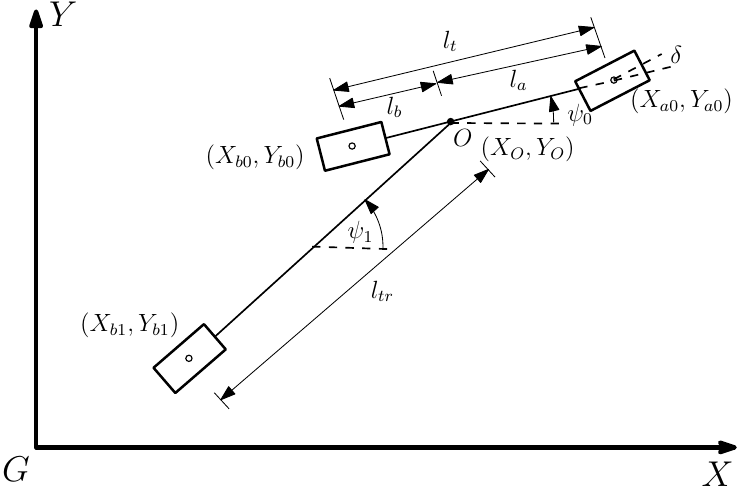}
	\caption{Schematic of single-track tractor-semitrailer combination.}
	\label{fig:st_model}
\end{figure}

Figure~\ref{fig:st_model} depicts the schematic of the tractor-semitrailer combination. The positions of the centre of axles in the global reference system ($G$) are given by:
\begin{itemize}
	\item Front axle of tractor: $(X_{a0}, Y_{a0})$,
	\item Rear axle of tractor: $(X_{b0}, Y_{b0})$,
	\item Rear axle of semitrailer: $(X_{b1}, Y_{b1})$,
\end{itemize}
where the second subscript in the position, `$0$' or `$1$', denotes the first (tractor) or second (semitrailer) units, respectively. The articulation joint, denoted by `O', is located at a distance of $l_a$ and $l_b$ from the tractor's front and rear axle, respectively. The rear axle of the semitrailer is located at $l_{tr}$ from the articulation joint. The yaw angle of the tractor is represented by $\psi_0$, whereas $\psi_1$ indicates the yaw angle of the semitrailer. Furthermore, $\delta$ denotes the steering angle of the first axle of the tractor.

The non-holonomic constraints for the tractor-semitrailer combination depicted in Figure~\ref{fig:st_model} are:
\begin{subequations}
	\begin{align}
		 & \dot{X}_{a0} \sin \: (\psi_{0} + \delta) - \dot{Y}_{a0} \cos \: (\psi_{0} + \delta) = 0, \label{eq: a0} \\
		 & \dot{X}_{b0} \sin \psi_0 - \dot{Y}_{b0} \cos \psi_0 = 0,                                                \\
		 & \dot{X}_{b1} \sin \psi_1 - \dot{Y}_{b1} \cos \psi_1 = 0. \label{eq: b1}
	\end{align}
	\label{eq: nonhol}
\end{subequations}
The following relations hold based on the kinematics of the combination.
\begin{subequations}
	\begin{align}
		X_{a0} & = X_{b0} + l_{t} \cos \psi_0,                    \\
		Y_{a0} & = Y_{b0} + l_{t} \sin \psi_0,                    \\
		X_{b1} & = X_{b0} + l_b \cos \psi_0 - l_{tr} \cos \psi_1, \\
		Y_{b1} & = Y_{b0} + l_b \sin \psi_0 - l_{tr} \sin \psi_1.
	\end{align}
	\label{eq: position}
\end{subequations}

Let $u_{b0}$ represent the longitudinal speed of the tractor's rear axle. Assuming a small side-slip angle at the rear axle, the lateral speed component $v_{b0}$ is considered negligible. Under this assumption, the dynamics associated with the rear axle of the tractor can be expressed as:
\begin{subequations}
	\begin{align}
		\dot{X}_{b0} & = u_{b0} \cos \psi_0 - v_{b0} \sin \psi_0 \approx u_{b0} \cos \psi_0, \label{eq:Xb0dot} \\
		\dot{Y}_{b0} & = u_{b0} \sin \psi_0 + v_{b0} \cos \psi_0 \approx u_{b0} \sin \psi_0.
	\end{align}
	\label{eq:XYb0dot}
\end{subequations}
By differentiating the position expressions in (\ref{eq: position}) and substituting (\ref{eq:XYb0dot}), the following relation is obtained:
\begin{subequations}
	\begin{align}
		\dot{X}_{a0} & = u_{b0} \cos \psi_0 - l_{t} \dot{\psi}_0 \sin \psi_0 ,                                \\
		\dot{Y}_{a0} & = u_{b0} \sin \psi_0 + l_{t} \dot{\psi}_0 \cos \psi_0 ,                                \\
		\dot{X}_{b1} & = u_{b0} \cos \psi_0 - l_b \sin \psi_0 \dot{\psi}_0 + l_{tr} \dot{\psi}_1 \sin \psi_1, \\
		\dot{Y}_{b1} & = u_{b0} \sin \psi_0 + l_b \cos \psi_0 \dot{\psi}_0 - l_{tr} \dot{\psi}_1 \cos \psi_1,
	\end{align}
	\label{eq: speed}
\end{subequations}
Now, the yaw rate of the tractor ($\dot{\psi}_0$) can be computed by substituting the expressions obtained from (\ref{eq: speed}) in (\ref{eq: a0}).
\begin{equation}
	\begin{aligned}
		                      & \dot{X}_{a0} \sin \: (\psi_{0} + \delta) - \dot{Y}_{a0} \cos \: (\psi_{0} + \delta) = 0,                                                         \\
		\Leftrightarrow \quad & (u_{b0} \cos \psi_0 - l_{t} \dot{\psi}_0 \sin \psi_0 )\sin \: (\psi_{0} + \delta) - \cdots                                                         \\
		                      &  \quad  (u_{b0} \sin \psi_0 + l_{t} \dot{\psi}_0 \cos \psi_0) \cos \: (\psi_{0} + \delta) = 0,                                 \\
		\Leftrightarrow \quad & \left(  \cos \psi_0\sin \: (\psi_{0} + \delta) -  \sin \psi_0 \cos \: (\psi_{0} + \delta) \right) u_{b0} - \cdots                                  \\
		                      &   \left(  \sin \psi_0 \sin(\psi_{0} + \delta) + \cos \psi_0 \cos \: (\psi_{0} + \delta) \right) \dot{\psi}_0  l_{t} = 0, \\
		\Leftrightarrow \quad & u_{b0} \sin \delta - l_{t} \dot{\psi}_{0} \cos \delta = 0,                                                                                       \\
		\Leftrightarrow \quad & \dot{\psi}_0 = \frac{u_{b0} \tan \delta }{l_{t}}.
	\end{aligned}
	\label{eq: psi0dot}
\end{equation}
The yaw rate of the semitrailer ($\dot{\psi}_1$) can be computed by substituting the expressions obtained from (\ref{eq: speed}) in (\ref{eq: b1}).
\begin{equation}
	\begin{aligned}
		                      & \dot{X}_{b1} \sin \psi_1 - \dot{Y}_{b1} \cos \psi_1 = 0,                                                                                                  \\
		\Leftrightarrow \quad & (u_{b0} \cos \psi_0 - l_b \sin \psi_0 \dot{\psi}_0 + l_{tr} \dot{\psi}_1 \sin \psi_1) \sin \psi_1 + \cdots                                                 \\
		                      &  ( l_{tr} \dot{\psi}_1 \cos \psi_1 - u_{b0} \sin \psi_0 - l_b \cos \psi_0 \dot{\psi}_0 ) \cos \psi_1 = 0,                          \\
		\Leftrightarrow \quad & u_{b0} \sin \: (\psi_{1} - \psi_{0}) - l_b \dot{\psi}_{0} \cos \: (\psi_{1} - \psi_{0}) + l_{tr} \dot{\psi}_{1} = 0,                                      \\
		\Leftrightarrow \quad & \dot{\psi}_{1} = \frac{l_b \dot{\psi}_{0} \cos \: (\psi_{1} - \psi_{0}) - u_{b0} \sin \: (\psi_{1} - \psi_{0})}{l_{tr}},                                  \\
		\Leftrightarrow \quad & \dot{\psi}_{1} = \frac{u_{b0} \tan \delta}{l_{t}} \frac{l_b}{l_{tr}} \cos \: (\psi_{1} - \psi_{0}) - \frac{u_{b0}}{l_{tr}} \sin \: (\psi_{1} - \psi_{0}).
	\end{aligned}
	\label{eq: psi1dot}
\end{equation}
In the case where the tractor is driving straight, the heading of the tractor is constant, i.e $\dot{\psi}_0 = 0$. For this case, (\ref{eq: psi1dot}) can further be simplified to:
\begin{equation}
	\dot{\psi}_{1} = - \frac{u_{b0}}{l_{tr}} \sin \: (\psi_{1} - \psi_{0} (t_0)),
	\label{eq: psi1dotSimplified}
\end{equation}
where $\psi_{0} (t_0)$ denotes the initial yaw angle of the tractor. An analytical expression for the yaw angle of the semitrailer can be computed by integrating the yaw rate obtained in  (\ref{eq: psi1dotSimplified}).

\begingroup
\small
\begin{equation}
\begin{aligned}
&\int_{\psi_1(t_0)}^{\psi_1(t)}
\frac{1}{\sin(\psi_1-\psi_0(t_0))}\,d\psi_1
=
\int_{t_0}^{t}
-\frac{u_{b0}}{l_{tr}}\,dt,
\\[2ex]
&\psi_1(t)= \psi_0(t_0)+ \cdots \\
&\begin{cases}
\begin{aligned}
& 2\arctan\!\Bigg(
\left|
\tan\frac{\psi_1(t_0)-\psi_0(t_0)}{2}
\right| \cdots\\
&\times
\exp\!\left(
\frac{u_{b0}}{l_{tr}} \,  (t_0-t)
\right)
\Bigg),
\end{aligned}
&
\begin{aligned}
&:\psi_1(t-1)-\psi_0(t_0)<0,
\end{aligned}
\\[2ex]
\begin{aligned}
& -2\arctan\!\Bigg(
\left|
\tan\frac{\psi_1(t_0)-\psi_0(t_0)}{2}
\right| \cdots\\
&\times
\exp\!\left(
\frac{u_{b0}}{l_{tr}} \,  (t_0-t)
\right)
\Bigg),
\end{aligned}
&
\begin{aligned}
&:\psi_1(t-1)-\psi_0(t_0)>0.
\end{aligned}
\end{cases}
\end{aligned}
\label{eq: psi1}
\end{equation}
\endgroup
The above integration is performed by assuming a constant speed ($u_{b0}$). When the speed is not constant, $u_{b0}$ is reformulated using (\ref{eq:Xb0dot}) and is given by
\begin{equation}
	u_{b0} =\dfrac{\dot{X}_{b0}}{\cos \, \psi_0}.
\end{equation}
Substituting the expression for $u_{b0}$ into \eqref{eq: psi1dotSimplified} and integrating yields the semitrailer's yaw angle for non-constant speed:
\begingroup
\small
\begin{equation}
\begin{aligned}
&\int_{\psi_1(t_0)}^{\psi_1(t)}
\frac{d\psi_1}{\sin(\psi_1-\psi_0(t_0))}
=
\int_{t_0}^{t}
-\frac{u_{b0}}{l_{tr}}\,dt,
\\
&\int_{\psi_1(t_0)}^{\psi_1(t)}
\frac{d\psi_1}{\sin(\psi_1-\psi_0(t_0))}
=
\int_{t_0}^{t}
-\frac{\dot X_{b0}}{l_{tr}\cos\psi_0}\,dt,
\\[1ex]
&\psi_1(t)=\psi_0(t_0)+ \cdots \\
&\begin{cases}
\begin{aligned}
& 2\arctan\!\Bigg(
\left|
\tan\frac{\psi_1(t_0)-\psi_0(t_0)}{2}
\right| \cdots\\
&\times
\exp\!\left(
\frac{X_{b0}(t_0)-X_{b0}(t)}
{l_{tr}\cos\psi_0}
\right)
\Bigg),
\end{aligned}
&
\begin{aligned}
&:\psi_1(t-1)-\psi_0(t_0)<0,
\end{aligned}
\\[2ex]
\begin{aligned}
& -2\arctan\!\Bigg(
\left|
\tan\frac{\psi_1(t_0)-\psi_0(t_0)}{2}
\right| \cdots\\
&\times
\exp\!\left(
\frac{X_{b0}(t_0)-X_{b0}(t)}
{l_{tr}\cos\psi_0}
\right)
\Bigg),
\end{aligned}
&
\begin{aligned}
&:\psi_1(t-1)-\psi_0(t_0)>0.
\end{aligned}
\end{cases}
\end{aligned}
\label{eq: psi1_acc}
\end{equation}
\endgroup



\end{document}